\RequirePackage{fix-cm} 
\documentclass[10pt]{article}

\usepackage{paper}
\usepackage{colortbl}

\usepackage{amsmath,amsfonts,bm}

\def\eqref#1{equation~\ref{#1}}

\def\1{\bm{1}}

\DeclareMathAlphabet{\mathsfit}{\encodingdefault}{\sfdefault}{m}{sl}
\SetMathAlphabet{\mathsfit}{bold}{\encodingdefault}{\sfdefault}{bx}{n}

\usepackage{amsmath}
\usepackage{amssymb}
\usepackage{booktabs}
\usepackage{float}
\usepackage{graphicx}
\usepackage{microtype}
\usepackage{multirow}
\usepackage{tabularx}
\usepackage{tikz}
\usepackage{url}
\usepackage{xcolor}
\usepackage{hyperref}
\usepackage{fvextra}
\usetikzlibrary{arrows.meta,calc,positioning}

\definecolor{contextblue}{HTML}{EAF2FF}
\definecolor{contextblueedge}{HTML}{4A78C2}
\definecolor{policygreen}{HTML}{ECF7EE}
\definecolor{policygreenedge}{HTML}{4F8A5B}
\definecolor{toolorange}{HTML}{FFF2E2}
\definecolor{toolorangeedge}{HTML}{C47A2C}
\definecolor{neutralgray}{HTML}{F4F5F7}
\definecolor{neutralgrayedge}{HTML}{737A86}

\definecolor{promptviolet}{HTML}{8256B5}
\definecolor{promptblue}{HTML}{2D78AE}
\definecolor{promptwarm}{HTML}{008F83}
\newcommand{\promptlabel}[1]{\textcolor{promptviolet}{#1}}
\newcommand{\promptkey}[1]{\textcolor{promptblue}{#1}}
\newcommand{\promptcue}[1]{\textcolor{promptwarm}{#1}}
\newtcolorbox{promptcard}[3][]{%
  enhanced,breakable,lines before break=8,colback=black!2,colframe=black!50,
  colbacktitle=black!50,coltitle=white,
  boxrule=0.5pt,arc=3pt,left=7pt,right=7pt,top=6pt,bottom=6pt,
  fonttitle=\sffamily\bfseries\small,title={#2},
  title after break={#2\ (continued)},
  fontupper=\normalfont\ttfamily\fontsize{8}{10.5}\selectfont\raggedright,
  before upper={\promptlabel{[#3]}\par\smallskip},
  before skip=8pt,after skip=8pt,
  #1
}
\newcommand{\promptpart}[1]{\par\smallskip\noindent\promptkey{[#1]}\par\nobreak}
\DefineVerbatimEnvironment{promptformat}{Verbatim}{%
  fontsize=\fontsize{8}{10}\selectfont,
  commandchars=\\\|\!,
  breaklines=true,breakanywhere=true,breaksymbolleft={},breaksymbolright={},
  baselinestretch=1,tabsize=2}

\newcommand{\method}{\textsc{GPT-Policy}}
\renewcommand{\PaperTitle}{In-Context Robot Learning with VLM Agents}
\renewcommand{\PaperAuthors}{Dongzhou~Cheng$^{1,2*}$, Taoran~Yi$^{3,1*}$, Ye~Fang$^{4,1*}$, Xingwu~Zhang$^{1,5*}$, Fan~Feng$^{1,6*}$, Yixuan~Li$^{1,6*}$, Gengxiong~Zhuang$^{1,10*}$, Rongze~Wang$^{1,4}$, Shuai~Yang$^{1,7}$, Wei~Song$^{2}$, Weizhi~Xue$^{1,8}$, Minyan~Wu$^{1}$, Jie~Gui$^{9}$, Jiaqi~Wang$^{2}$, Tong~Wu$^{1,4\dagger}$}
\renewcommand{\PaperAffiliation}{\small
  \mbox{$^{1}$Morphi Robot}\quad
  \mbox{$^{2}$Shanghai Innovation Institute}\quad
  \mbox{$^{3}$Huazhong University of Science and Technology}\quad
  \mbox{$^{4}$Fudan University}\quad
  \mbox{$^{5}$Hunan University}\quad
  \mbox{$^{6}$The Chinese University of Hong Kong}\quad
  \mbox{$^{7}$Shanghai Jiao Tong University}\quad
  \mbox{$^{8}$Wuhan University}\quad
  \mbox{$^{9}$Southeast University}\quad
  \mbox{$^{10}$Beihang University}}
\renewcommand{\PaperContributions}{%
  \vspace{3pt}%
  {\sffamily\footnotesize\color{paperink!75}%
    $^{*}$\enspace Co-first authors.\qquad $^{\dagger}$\enspace Corresponding authors.}}
\renewcommand{\PaperMetadata}{}
\renewcommand{\BrandLogoPath}{}
\renewcommand{\MarginMark}{}

\renewcommand{\PaperAbstract}{%

Enabling robots to adapt to unfamiliar environments as readily as humans remains a moonshot goal of embodied AI. No finite collection of demonstrations can cover every task and situation a robot will encounter, making the ability to learn from context at deployment essential for generalization. Such in-context learning (ICL), however, remains largely beyond the reach of existing robotic policies. The broad agentic capabilities of commercial vision-language models (VLMs), such as GPT-6 Astra, raise a compelling question: can these models learn from demonstrations, examples, and interaction feedback, then translate that information into executable and verifiable robot behavior from a new initial state without gradient updates or persistent changes to task-specific parameters? We introduce {\bfseries GPT-Policy}, a general-agent framework for in-context robot learning. GPT-Policy integrates a context compiler that preserves task-relevant visual transitions, a VLM that proposes robot-tool actions, and a constrained controller that verifies and executes each action and reports its outcome. We evaluate its reliability and limitations through task success and efficiency metrics, matched comparisons across models, and controlled context ablations. In real-robot trials, human video demonstrations improve task completion even without robot action labels, while aligned action references yield further gains on contact-sensitive tasks. These findings position GPT-Policy as a step toward robot adaptation through in-context learning, providing an empirical foundation for translating the general-purpose capabilities of VLMs into physical behavior and clarifying the challenges that must be overcome for reliable deployment.\par\medskip{\sffamily\bfseries\urlstyle{tt}\color{black}\noindent Code: \href{https://github.com/cheng-haha/GPT-Policy}{\textcolor[rgb]{0.035,0.412,0.855}{\mdseries\nolinkurl{https://github.com/cheng-haha/GPT-Policy}}}\par\noindent Website: \href{https://cheng-haha.github.io/GPT-Policy/}{\textcolor[rgb]{0.035,0.412,0.855}{\mdseries\nolinkurl{https://cheng-haha.github.io/GPT-Policy/}}}\par}

}


\begin{document}
\papermaketitle
\section{Introduction}


No training dataset can cover every situation a robot will encounter. Adapting to new tasks, unfamiliar object arrangements, and unexpected interactions during deployment is therefore a central challenge for embodied AI~\citep{brohan2023rt2,kim2024openvla,octo2024octo,black2025pi0}. Humans routinely adapt to such situations by observing others, interpreting examples, and learning from the consequences of their actions. Enabling robots to learn in this way through in-context learning (ICL) is a key step toward general embodied agents~\citep{duan2017oneshot,fu2024icrt,vosylius2024instant}. General-purpose language and vision-language models (VLMs) offer a promising starting point: their ability to learn from context suggests that some capabilities needed for robot adaptation may already be present without dedicated policy training~\citep{brown2020language,alayrac2022flamingo,huang2022zeroshot}. This raises a central question: to what extent can these models use contextual information to guide robot behavior in unfamiliar situations?

\begin{figure}[!t]
    \centering
    \includegraphics[width=\linewidth]{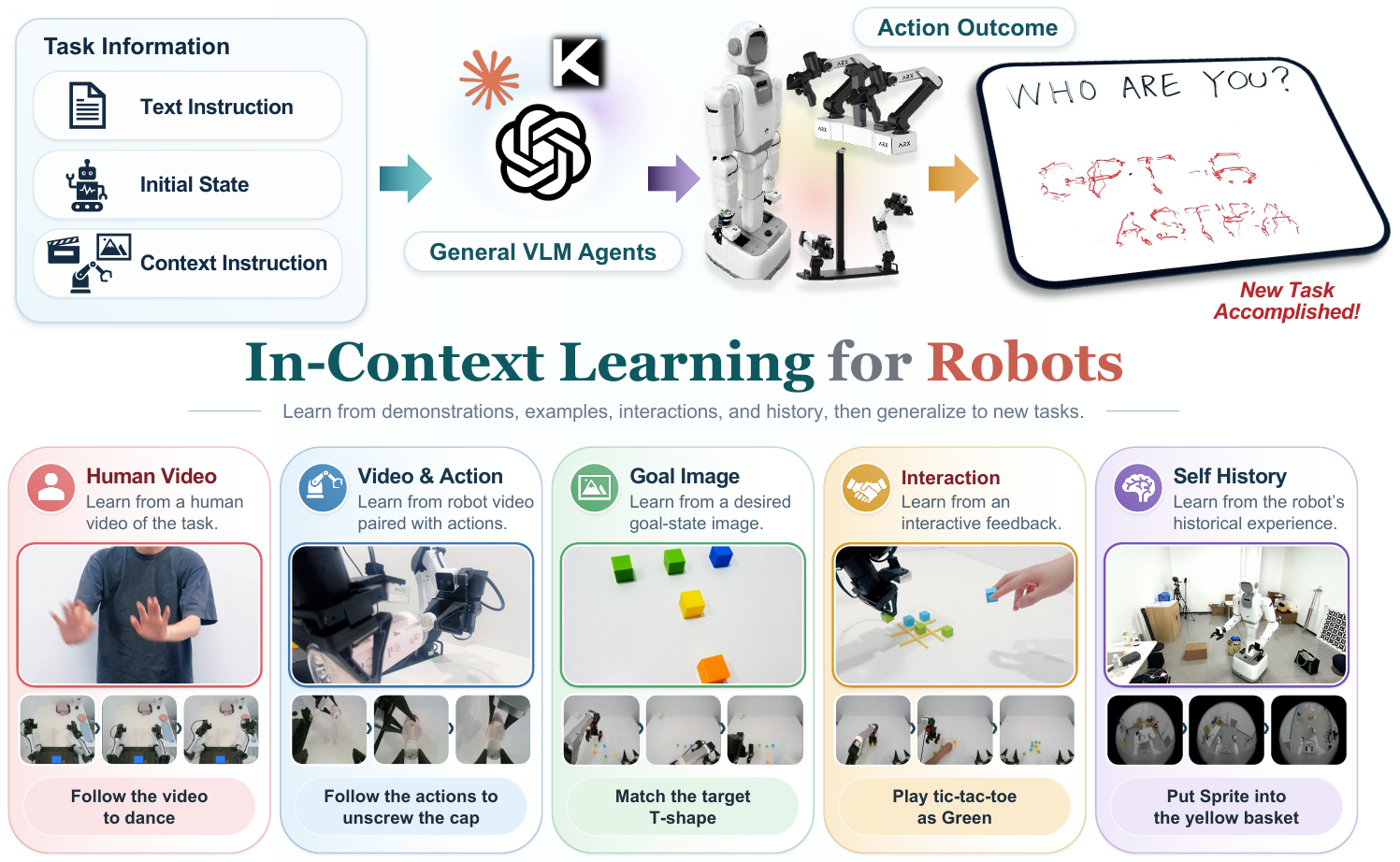}
    \caption{\textbf{In-context robot control with GPT-Policy.}
    A general-purpose VLM combines the task instruction, initial state, and
    contextual information to guide robot actions. Context can include human
    videos, robot videos with recorded actions, goal images, human--robot
    interaction, and self-interaction history. The examples illustrate
    manipulation, interactive play, and mobile object retrieval.}
    \label{fig:teaser}
\end{figure}

We define robotic ICL as the ability to adapt behavior based on demonstrations, examples, or interaction experience provided at test time, without gradient updates or persistent task-specific parameter changes. This requires a robot to extract relevant information from context and apply it to its current situation, even when that situation differs from the demonstrations. Each form of context can guide a different aspect of behavior: goal images specify desired outcomes, human and robot videos illustrate procedures, and aligned robot actions provide motion references. Interaction history records previous observations and outcomes, while online human feedback clarifies intent or changes in environmental rules. The central challenge is to determine which information matters for the current decision and translate it into appropriate physical action.


Recent robot foundation models have begun to demonstrate this capability. GEN-1.5 reports one-shot skill adaptation from physical prompts, including human-to-robot and sim-to-real examples~\citep{generalist2026gen15}. S1 uses video demonstrations to specify novel atomic and long-horizon tasks~\citep{skild2026s1}, while Zero-WAM trains a video-action model to follow human video guidance on unseen tasks~\citep{zhou2026zerowam}. These advances motivate a complementary question: \emph{to what extent can off-the-shelf, general-purpose VLMs support robotic in-context learning without being trained as dedicated robot policies?} Answering this question can help distinguish the task understanding available in general models from the capabilities that require specialized embodied learning.

To investigate this question, we introduce \method, a general-agent framework that connects an off-the-shelf VLM to robot tools through a shared closed-loop interface. A context compiler preserves task-relevant visual transitions and available action references. The VLM interprets this context alongside the current scene and proposes parameterized robot-tool actions. A constrained execution layer checks and executes the proposed actions, then returns observations and outcomes to support replanning. This interface allows us to examine how different models use different forms of context within a common execution framework.



Our evaluation covers \textit{five context families} spanning cross-embodiment imitation, contact-sensitive manipulation, goal-image following, active exploration, and human-robot interaction. Through matched model comparisons and controlled context ablations, we measure whether context improves task completion and how it changes decision count and execution time. The results show that task-relevant context can improve success while reducing decisions and execution time. In real-robot trials, human videos improve task completion even without robot action labels, while aligned action references provide further gains on contact-sensitive tasks. Yet the same experiments expose a consequential gap: better task understanding and action selection do not ensure precise contact, reliable outcome verification, or physical safety. Context can guide a robot toward the right behavior while leaving critical execution failures unresolved.



These findings make robotic ICL a concrete question about where adaptation succeeds and where it breaks down across the perception–action loop. General-purpose VLMs can already use heterogeneous context to inform robot decisions, providing a starting point for adaptation beyond the training distribution. The next challenge is to make that adaptability dependable throughout physical execution. By providing a common framework and empirical evidence for studying this gap, \method~helps define a research agenda for embodied foundation models and identifies several promising directions for near-term research.

\section{Related Work}

\paragraph{General-purpose agents for robot control.}
Recent general-purpose models, including GPT-6 Astra, Claude Fable, Kimi K3, and GLM-5.3, support reasoning, tool use, and multi-step task execution~\citep{openai2026astra,anthropic2026fable,moonshot2026kimi,zai2026glm}. In robotics, language and vision--language models have been applied to affordance-grounded skill selection~\citep{ahn2022saycan}, program synthesis over perception and control APIs~\citep{liang2023code}, and spatial objective construction for motion planning~\citep{huang2023voxposer}. Agentic systems further automate policy development through execution feedback: ASPIRE diagnoses program failures and distills validated repairs into reusable skills~\citep{lu2026aspire}, while ENPIRE enables coding agents to refine robot policies and training procedures through repeated real-world trials~\citep{xiao2026enpire}. Closer to online action selection, RoboPrompt predicts robot actions from textual prompts encoding object poses and expert end-effector actions~\citep{yin2025roboprompt}. Show-Harness enables closed-loop VLM control through a discrete semantic action interface and uses video demonstrations to condition task planning~\citep{chen2026showharness}. Our study focuses on how context shapes the embodied behavior of a general-purpose agent.

\paragraph{In-context learning for robot policies.}
Demonstration-conditioned robot control builds on in-context learning~\citep{brown2020language} and one-shot imitation~\citep{duan2017oneshot}, using examples provided at deployment to specify the desired behavior. Existing policies extract task information from robot sensorimotor trajectories~\citep{fu2024icrt,sridhar2025ricl,generalist2026gen15} or geometric demonstration representations~\citep{vosylius2024instant}. This paradigm also accommodates human visual demonstrations, allowing observed human behavior to guide robot execution~\citep{mimicdroid2026,patel2026behavior,zhou2026zerowam,skild2026s1}. Beyond individual demonstrations, RoboTTT investigates adaptation from long visual contexts that combine human demonstrations with robot interaction history~\citep{jiang2026robott}. Despite differences in context representation and adaptation mechanism, these works share a focus on enabling robot policies and embodied foundation models to exploit demonstration context. Our study instead examines this capability in an off-the-shelf general-purpose multimodal agent: how visual demonstrations and interaction history inform its task interpretation and action selection, without additional robot-specific training or test-time parameter updates.

\begingroup
\setlength{\abovedisplayskip}{5pt}
\setlength{\belowdisplayskip}{5pt}
\setlength{\abovedisplayshortskip}{3pt}
\setlength{\belowdisplayshortskip}{3pt}
\section{Method}\label{sec:method}

GPT-Policy connects a general-purpose vision-language model (VLM) to robot tools through a shared context-to-action interface (Figure~\ref{fig:overview}). Embodiment-specific adapters translate tool requests into executable commands. We describe the task formulation, context construction, and closed-loop execution below.

\begin{figure}[!t]
    \centering
    \includegraphics[width=0.985\linewidth]
        {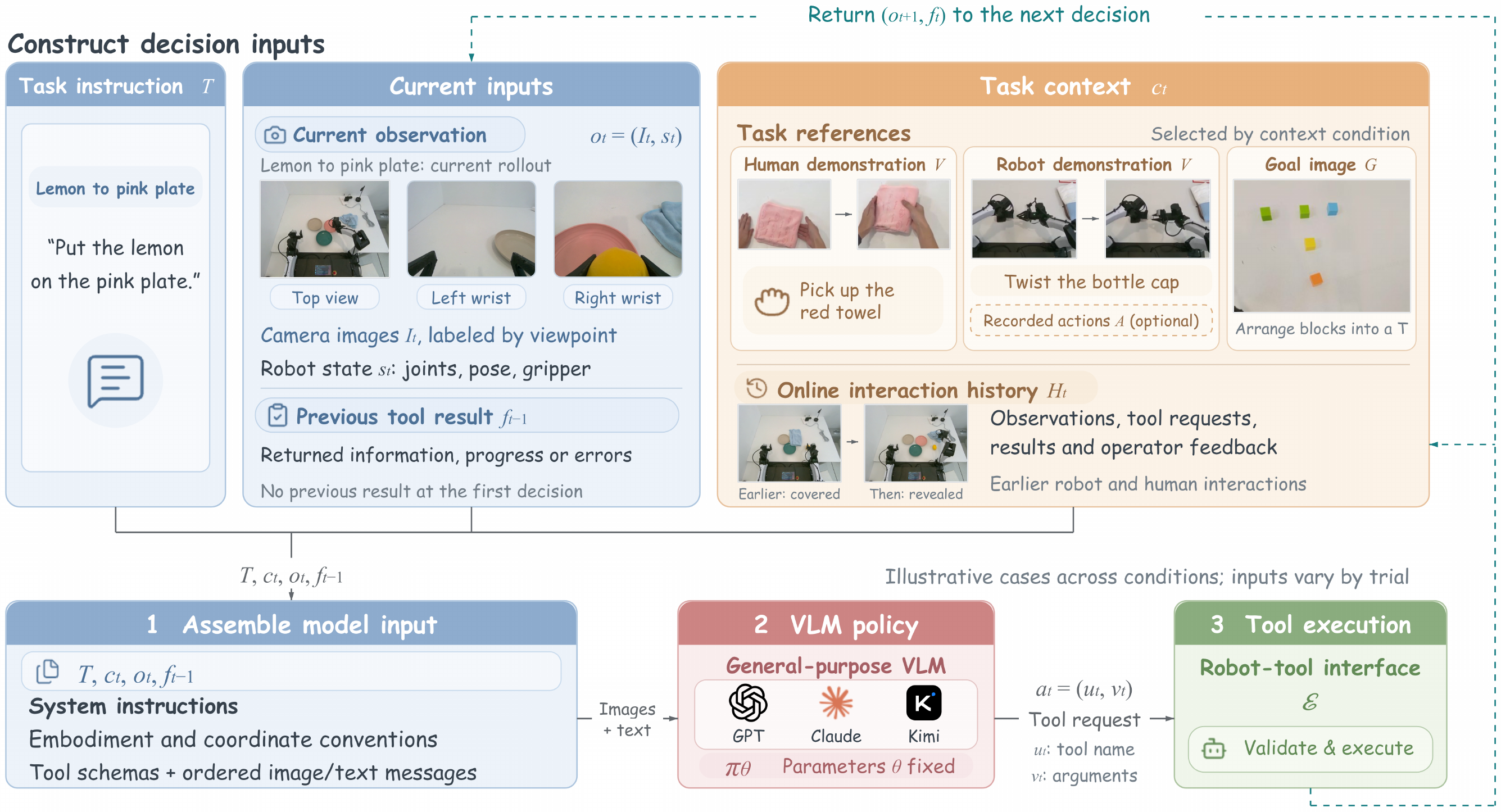}
    \caption{\textbf{Overall architecture of GPT-Policy.}
    The task instruction, current state, task references, and interaction
    history are combined with shared instructions and tool schemas to form
    the VLM input. A fixed VLM selects tool requests, which constrained
    robot tools execute. Returned observations and execution feedback
    update the context for the next decision. The images illustrate
    inputs from different experimental conditions.}
    \label{fig:overview}
\end{figure}

\subsection{Problem Formulation}\label{sec:problem-formulation}

Let $T$ denote the task instruction and $o_t=(I_t,s_t)$ the latest observation, where $I_t$ comprises images labeled by camera view and $s_t$ denotes the robot state. The context $c_t$ contains the task references and online interaction history available at decision step $t$. Depending on the context condition, it includes a goal image $G$, demonstration videos $V$ represented by selected keyframes, recorded action sequences $A$ with any accompanying measured robot states, and online interaction history $H_t$. System instructions define the embodiment, coordinate conventions, and tool schemas.

At step $t$, the model selects a tool request $a_t=(u_t,v_t)$, comprising a tool name $u_t$ and arguments $v_t$, conditioned on $T$, $c_t$, $o_t$, and the preceding tool result $f_{t-1}$:
\begin{equation}
\begin{gathered}
a_t \sim \pi_\theta(\,\cdot\mid T,c_t,o_t,f_{t-1}),\\
(o_{t+1},f_t)=\mathcal{E}(a_t,o_t).
\end{gathered}
\end{equation}
Here, $\pi_\theta$ is the VLM policy, whose parameters $\theta$ remain fixed during task execution, and $\mathcal{E}$ is the robot-tool interface. The result $f_t$ contains returned information, execution progress, or errors. After a completed or rejected request, the next decision receives updated observations and the preceding tool result. No preceding tool result is available at the first decision.

\subsection{Context Construction}\label{sec:context-construction}

\paragraph{Task references.}

Task references specify a desired outcome or illustrate a procedure. A $\langle\textit{goal image}\ G\rangle$ provides a visual reference for the target object arrangement or task outcome without prescribing intermediate actions. $\langle\textit{Demonstration videos}\ V\rangle$ show object interactions and action order in \emph{human demonstrations}, where a person performs the task, or \emph{teleoperated robot demonstrations}, where a human operator controls the robot. $\langle\textit{Recorded actions}\ A\rangle$ optionally supplement these videos and may include accompanying measured robot states. These reference records remain distinct from the model's tool requests $a_t$ in the current trial.

For model input, $V$ is encoded as timestamp-ordered frames paired with viewpoint identifiers, available annotations, and corresponding state or action records. The loader interleaves these text records with image blocks. Because annotations may describe the procedure, comparisons of recorded-action inputs must hold the selected images and non-action text fixed and specify the state information provided.

\paragraph{Online interaction history.}

During execution, $H_t$ records observations, tool requests, results, and operator feedback separately from the task references. Provider adapters manage this history by retaining reference inputs, limiting older live images, and either retaining accumulated text or replacing older exchanges with host-generated summaries. These updates preserve the ongoing robot trial and decision count.

\begin{figure}[!t]
\centering
\includegraphics[width=0.92\linewidth]{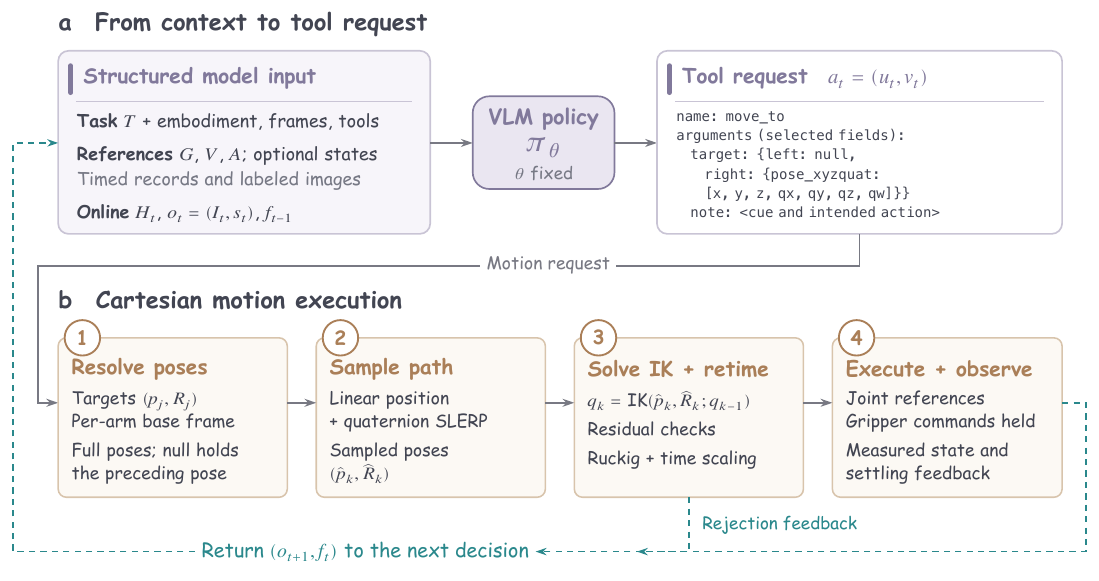}

\caption{\textbf{Details of the GPT-Policy execution harness.} (a) The VLM policy $\pi_\theta$ generates tool requests from interleaved image and text inputs; selected arguments are shown schematically. (b) The Cartesian adapter resolves targets, samples the pose path, checks IK residuals, times joint references, and executes the motion. Cartesian moves hold gripper commands; separate tools change the gripper opening. Solid arrows trace the forward path; dashed arrows return observations and execution or rejection feedback for the next decision.}
\label{fig:context-format}
\end{figure}

\subsection{Model Interface and Closed-Loop Execution}\label{sec:action-execution}\label{sec:execution}

Figure~\ref{fig:context-format} traces the closed loop from VLM tool requests to robot execution and feedback. The Cartesian adapter samples the requested pose path, solves inverse kinematics (IK), and assigns timestamps to the joint references. Robot observations and execution feedback inform the next VLM decision.

\paragraph{Structured tool interface.}

Inputs to $\pi_\theta$ interleave source- and view-labeled image blocks with text records for $T$, $c_t$, $o_t$, and $f_{t-1}$, together with system instructions and tool schemas. Provider adapters normalize a structured JSON selection or native tool call into \texttt{name} ($u_t$) and an \texttt{arguments} object ($v_t$). Figure~\ref{fig:context-format} illustrates the selected arguments of a motion request, including a \texttt{note} describing the observed cue and intended action. Appendix~\ref{app:prompt-details} provides the prompt templates, context formats, and representative task instructions.

Cartesian requests specify one target (\texttt{move\_to}) or an ordered sequence (\texttt{move\_eef\_chunk}) for the tool center point (TCP), a calibrated reference frame on the end effector. Each non-null target contains its position $p\in\mathbb{R}^3$ and orientation $R\in SO(3)$ as \texttt{pose\_xyzquat}, using the arm's base frame and $xyzw$ quaternion order. In bimanual requests, a \texttt{null} arm entry holds its preceding pose. Gripper commands remain fixed during Cartesian motion and change through \texttt{set\_gripper}.

\paragraph{Pose interpolation.}

Starting from $(p_0,R_0)$, the adapter connects successive targets $(p_j,R_j)$, $j=1,\ldots,J$, using linear position interpolation and quaternion spherical linear interpolation (SLERP) along the shorter rotation arc~\citep{shoemake1985animating}. For segment $j=0,\ldots,J-1$ and progress $s\in[0,1]$, the path is
\begin{equation}
\begin{gathered}
p_j(s)=(1-s)p_j+s p_{j+1},\\
R_j(s)=R_j\exp\!\left(s\operatorname{Log}(R_j^{\top}R_{j+1})\right).
\end{gathered}
\end{equation}
The adapter samples this geometric path before solving IK.

\paragraph{Inverse kinematics.}

After converting each sampled TCP pose to the backend's kinematic frame, IK computes a joint reference from the preceding solution, initialized with the measured joint configuration at planning time:
\begin{equation}
q_k=\operatorname{IK}(\widehat p_k,\widehat R_k;q_{k-1}).
\end{equation}
The shared residual requirements are
\begin{equation}
\lVert e_{p,k}\rVert_2\leq\epsilon_p,\qquad
\lVert e_{R,k}\rVert_2\leq\epsilon_R.
\label{eq:ik-acceptance}
\end{equation}
Here $e_{p,k}$ and $e_{R,k}$ are the position and orientation residuals, and $\epsilon_p,\epsilon_R$ are their execution tolerances. Numerical stopping criteria and joint-bound handling depend on the IK backend; Appendix~\ref{app:motion-details} specifies these distinctions. Sequential seeding does not impose a hard bound on the joint displacement between samples.

\paragraph{Execution and feedback.}

After IK, Ruckig~\citep{berscheid2021jerk} assigns timestamps through a scalar progress profile. Time scaling then enforces the sampled joint velocity, acceleration, and jerk limits (Appendix~\ref{app:motion-details}). The backend dispatches the timed joint references and reports measured state, endpoint errors, and settling status. Completed and rejected requests supply feedback for the next decision; faults or operator interruption can end the trial. Model-declared completion remains distinct from physical task success.

Configuration-specific extensions can add geometric observations, motion rejection rules, or a separate completion review that returns unverified completion requests to the control loop.
\par
\endgroup

\section{Experiments}

This section evaluates whether a general-purpose agent can adapt to tasks during closed-loop robotic-arm operation by using different forms of context. The experiments are designed to answer three questions:

1. What information does each form of context provide?

2. Can this information improve task completion rates and reduce the number of decisions?

3. Which forms of context remain effective for fine contact-rich manipulation, deformable-object handling, and human-robot collaboration?

\subsection{Experimental Setup}

Each episode starts from a reset scene and ends when the task succeeds, the execution budget is exhausted, or a safety termination condition is triggered. An episode is counted as successful only when the final scene satisfies the task-specific geometric and semantic success criteria. All conditions use the same success criteria and termination rules. Each experiment is repeated three times.

The main metrics are:

\begin{itemize}
\item \textbf{Success Rate (S/T)}: \(S\) denotes the number of successful trials and \(T\) denotes the total number of trials; the metric is reported as \(S/T\).
\item \textbf{Decisions}: The number of decisions generated by the agent in one episode. Under the task-specific counting convention, each generated action target or action block counts as one decision.
\item \textbf{Execution Time}: The total time required to complete the task.
\end{itemize}

Robot, sensing, planning, and execution configurations are summarized in Appendix~\ref{app:configurations}. The experimental results are summarized in Table~\ref{tab:context-summary}.

\begingroup
\definecolor{TInk}{HTML}{25313B}
\definecolor{TMuted}{HTML}{5E6872}
\definecolor{TLine}{HTML}{D8E0E7}
\definecolor{TBlue}{HTML}{365D7D}
\definecolor{THuman}{HTML}{E5F0FA}
\definecolor{TRobot}{HTML}{E7F3ED}
\definecolor{TImage}{HTML}{FCF1DE}
\definecolor{THistory}{HTML}{F0EAF8}
\definecolor{THRI}{HTML}{F9EBE5}

\newcommand{\TSuccess}[1]{#1\,/\,3}
\newcommand{\TSection}[2]{%
  \multicolumn{5}{@{}l@{}}{%
    \begingroup\setlength{\fboxsep}{0pt}%
    \colorbox{#1}{\makebox[\linewidth][l]{%
      \hspace{7pt}\rule[-4.5pt]{0pt}{17pt}%
      {\sffamily\bfseries\fontsize{10}{11}\selectfont #2}%
    }}\endgroup%
  }\\[1.5pt]%
}
\newcommand{\TTask}[2]{\multirow{#1}{=}{\raggedright #2}}
\newcommand{\TCondition}[1]{#1}
\newcommand{\TBest}[1]{{\rmfamily\bfseries #1}}
\newcommand{\THeader}[1]{{\sffamily\bfseries\fontsize{9.6}{11}\selectfont#1}}

\begin{table}[!t]
\centering
\begingroup
\color{TInk}
\captionsetup{format=plain,labelfont={sf,bf,color=TBlue},font=normalsize,
  labelsep=period,justification=raggedright,singlelinecheck=false,skip=4pt}
\caption{\textbf{Results across context conditions.} GPT-6 Astra is evaluated on real robots across different context conditions. S/T denotes successful/total trials; bold entries mark the best-performing condition for each task. Decisions and time are averaged over all trials, including failures.}
\label{tab:context-summary}
\vspace{4pt}
\rmfamily\mdseries\fontsize{9}{11.5}\selectfont
\setlength{\tabcolsep}{5pt}
\setlength{\parskip}{0pt}
\arrayrulecolor{TLine}
\renewcommand{\arraystretch}{1.25}
\begin{tabularx}{\linewidth}{@{\hspace{8pt}}>{\raggedright\arraybackslash}p{119pt}>{\raggedright\arraybackslash}X>{\centering\arraybackslash}p{45pt}>{\centering\arraybackslash}p{54pt}>{\centering\arraybackslash}p{55pt}@{\hspace{8pt}}}
\toprule[0.8pt]
\THeader{Task} & \THeader{Context provided} & \THeader{S/T} & \THeader{Decisions} & \THeader{Time (min)} \\
\midrule[0.5pt]\noalign{\vskip2pt}
\TSection{THuman}{Human video demonstration}
\TTask{2}{Pick Red Towel} & \TCondition{None} & \TSuccess{0} & 96.3 & 24.6 \\
 & \TBest{\TCondition{Human Video}} & \TBest{\TSuccess{2}} & \TBest{76.7} & \TBest{18.9} \\
\noalign{\vskip1pt}\cmidrule[0.25pt](lr){1-5}
\TTask{2}{Pick Up Notebook} & \TCondition{None} & \TSuccess{0} & 94.0 & 24.6 \\
 & \TBest{\TCondition{Human Video}} & \TBest{\TSuccess{2}} & \TBest{66.7} & \TBest{16.1} \\
\noalign{\vskip3pt}
\TSection{TRobot}{Robot visual demonstration}
\TTask{3}{Unscrew Bottle Cap} & \TCondition{None} & \TSuccess{0} & 71.0 & 16.1 \\
 & \TCondition{Robot Video} & \TSuccess{2} & 74.3 & 15.2 \\
 & \TBest{\TCondition{Robot Video + Action}} & \TBest{\TSuccess{3}} & \TBest{54.7} & \TBest{17.9} \\
\noalign{\vskip1pt}\cmidrule[0.25pt](lr){1-5}
\TTask{3}{Remove and Reinsert Plug} & \TCondition{None} & \TSuccess{0} & 24.0 & 5.3 \\
 & \TCondition{Robot Video} & \TSuccess{0} & 33.7 & 7.9 \\
 & \TBest{\TCondition{Robot Video + Action}} & \TBest{\TSuccess{2}} & \TBest{48.3} & \TBest{10.8} \\
\noalign{\vskip3pt}
\TSection{TImage}{Target image}
Arrange T Shape & \TBest{\TCondition{Target Image}} & \TBest{\TSuccess{3}} & \TBest{66.7} & \TBest{15.8} \\
\noalign{\vskip1pt}\cmidrule[0.25pt](lr){1-5}
Arrange Fruit & \TBest{\TCondition{Target Image}} & \TBest{\TSuccess{3}} & \TBest{49.0} & \TBest{12.4} \\
\noalign{\vskip3pt}
\TSection{THistory}{Self-interaction history}
Lemon To Pink Plate & \TBest{\TCondition{Self History}} & \TBest{\TSuccess{3}} & \TBest{35.3} & \TBest{8.1} \\
\noalign{\vskip1pt}\cmidrule[0.25pt](lr){1-5}
Movable Exploration & \TBest{\TCondition{Self History}} & \TBest{\TSuccess{3}} & \TBest{40.33} & \TBest{25.53} \\
\noalign{\vskip3pt}
\TSection{THRI}{Online human-robot interaction}
Tic-Tac-Toe & \TBest{\TCondition{Human-Robot Interaction}} & \TBest{\TSuccess{3}} & \TBest{69.7} & \TBest{13.6} \\
\noalign{\vskip1pt}\cmidrule[0.25pt](lr){1-5}
Pointed Fruit Pickup & \TBest{\TCondition{Human-Robot Interaction}} & \TBest{\TSuccess{3}} & \TBest{67.3} & \TBest{15.0} \\
\bottomrule[0.5pt]
\end{tabularx}\par
\arrayrulecolor{black}
\endgroup
\end{table}
\endgroup

\begin{figure}[!t]
\centering
\includegraphics[width=\linewidth]{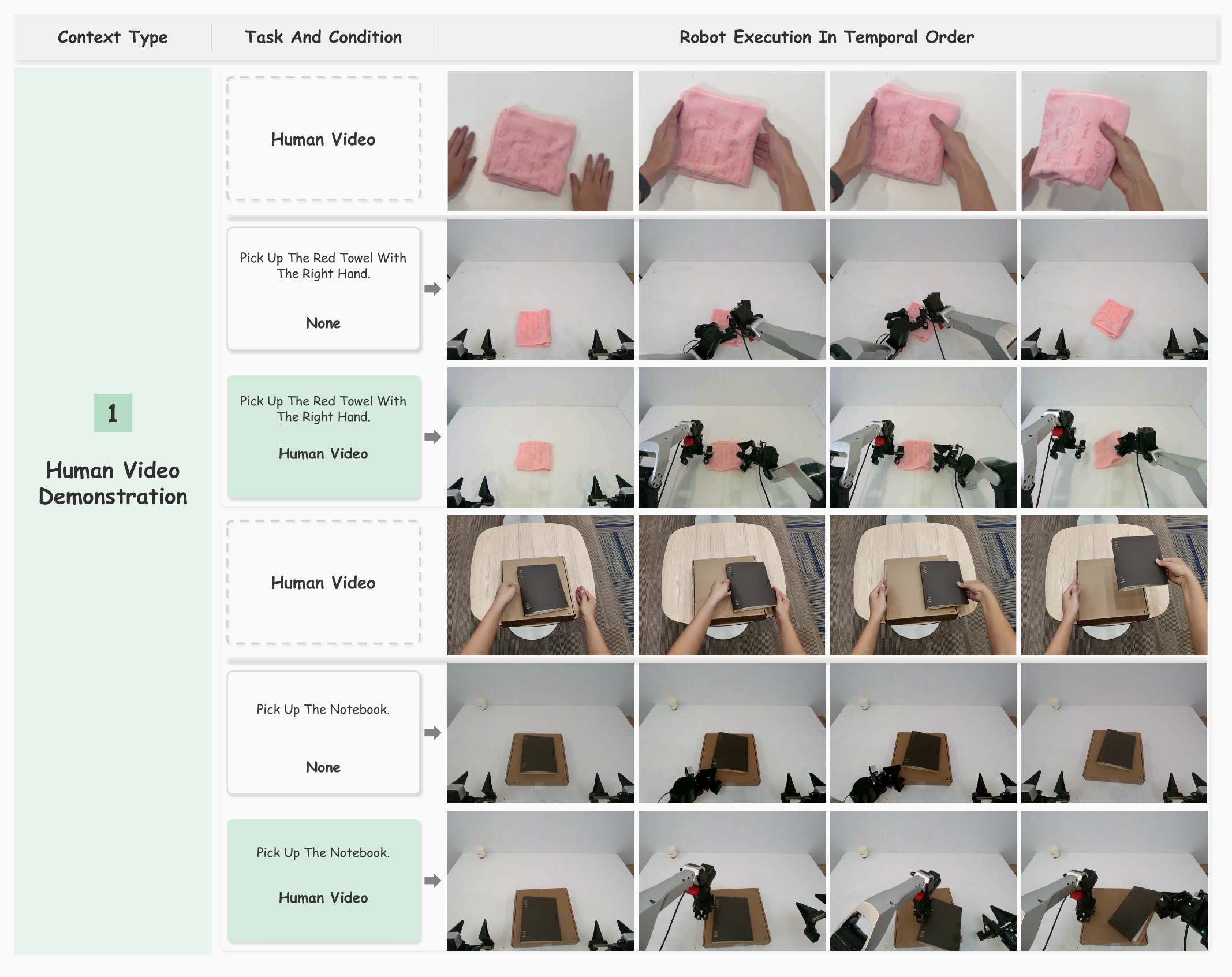}
\caption{\textbf{Human demonstrations and robot executions for towel and notebook pickup.} For each task, we show example runs under two context conditions: None and Human Video. Frames progress from left to right.}
\label{fig:human-video}
\end{figure}

\begin{figure}[!t]
\centering
\includegraphics[width=\linewidth]{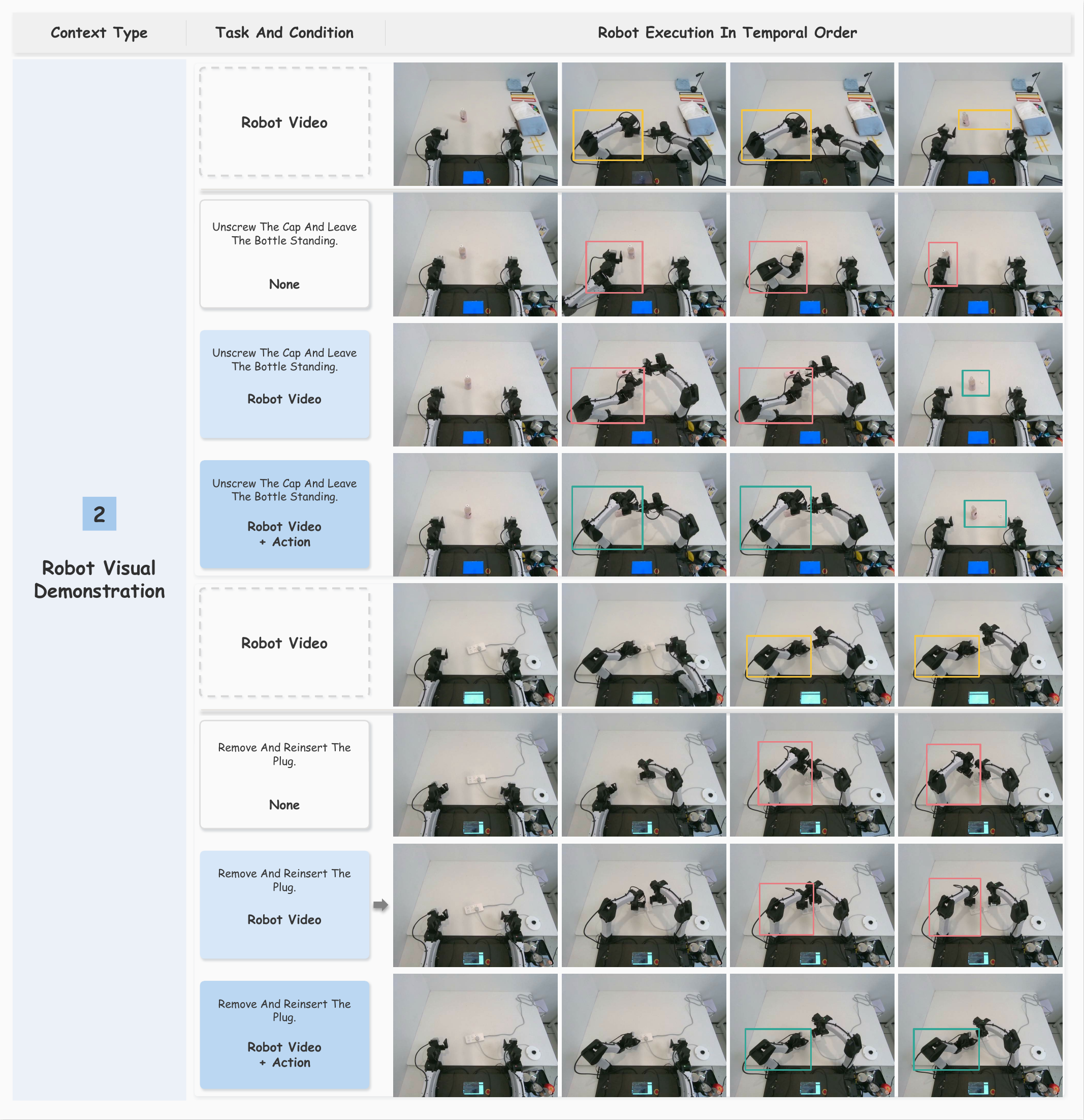}
\caption{\textbf{Robot demonstrations and executions for bottle opening and plug reinsertion.} For each task, we show example runs under three context conditions: None, Robot Video, and Robot Video + Action. Frames progress from left to right. Gold boxes mark reference regions in the demonstrations. Red marks local deviations; green marks closer matches.}
\label{fig:robot-visual}
\end{figure}

\begin{figure}[!t]
\centering
\includegraphics[width=\linewidth]{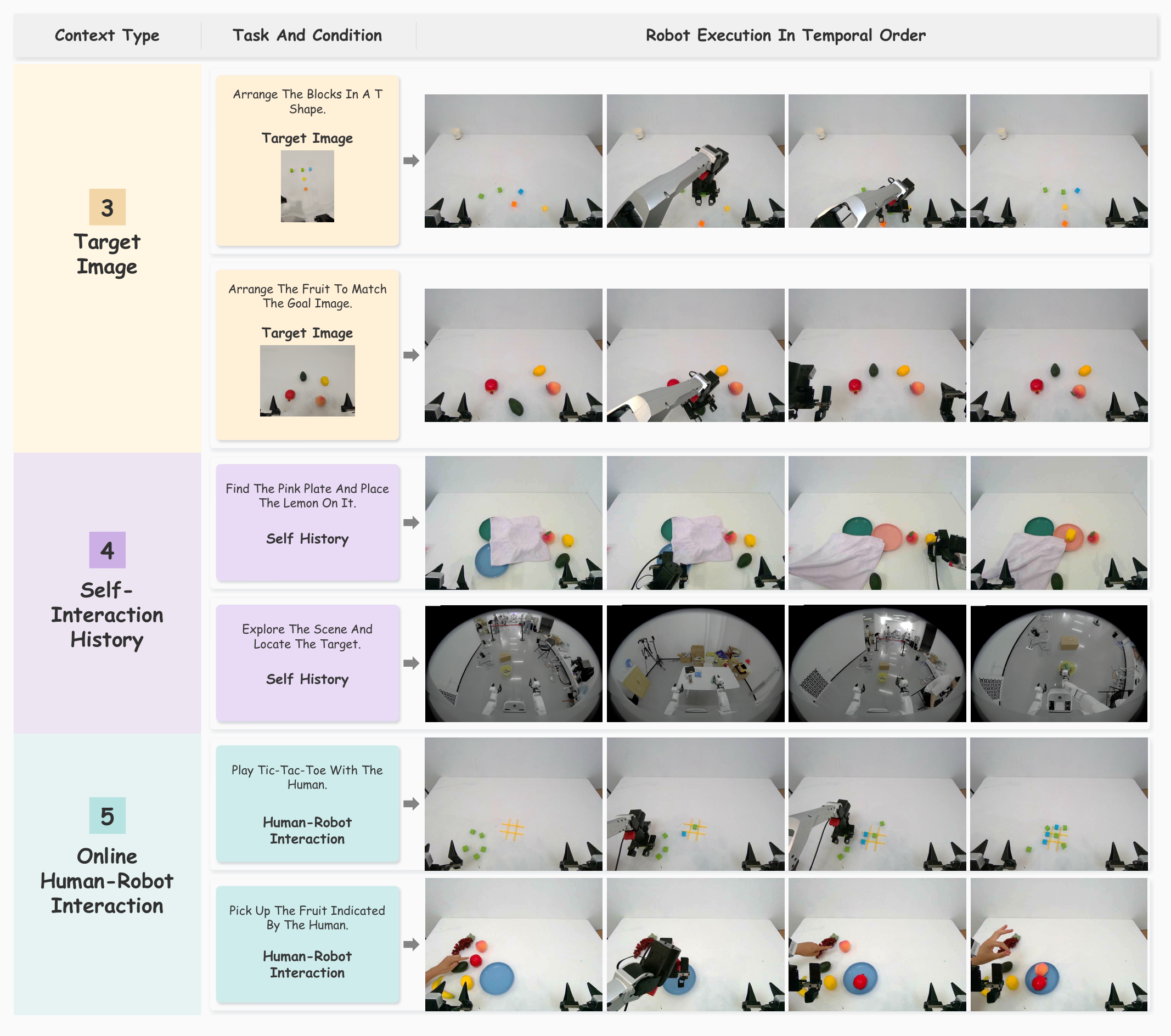}
\caption{\textbf{Goal images, self-interaction history, and online human interaction.} We show two example runs per context condition, one per row. Frames progress from left to right.}
\label{fig:other-contexts}
\end{figure}

\subsection{In-Context Learning from Human Videos}

We evaluate whether a single human demonstration video can guide GPT-6 Astra on ``Pick Red Towel'' and ``Pick Up Notebook.'' Human Video adds the video to the standard inputs; None uses the same instruction and observations without the Human Video. As shown in Table~\ref{tab:context-summary}, Human Video achieves 2/3 success on both tasks, compared with 0/3 under None. For towel pickup, the average decision count decreases from 96.3 to 76.7 and average execution time from 24.6 to 18.9 minutes; for notebook pickup, the corresponding averages decrease from 94.0 to 66.7 decisions and from 24.6 to 16.1 minutes.

The combination of higher success and lower average execution costs suggests that the demonstration provides useful procedural guidance. Figure~\ref{fig:human-video} shows grasping methods and interaction sequences that may help constrain the agent's choice of strategy. The human demonstration supplies no robot action labels; the agent generates robot-specific motion targets from current observations. These results are consistent with transferring an interaction strategy across embodiments.

\subsection{In-Context Learning from Robot Video and Actions}

\begin{samepage}
We compare None (no demonstration), Robot Video, and Robot Video + Action, which adds time-aligned end-effector poses, gripper states, and action commands to the same video. ``Unscrew Bottle Cap'' requires leaving the opened bottle standing securely; ``Remove and Reinsert Plug'' requires removing the plug and reinserting it into its original socket so that it remains fully seated after gripper release. In this condition order, Table~\ref{tab:context-summary} reports bottle-opening success of 0/3, 2/3, and 3/3, averaging 71.0, 74.3, and 54.7 decisions and 16.1, 15.2, and 17.9 minutes. Plug-task success is 0/3, 0/3, and 2/3, averaging 24.0, 33.7, and 48.3 decisions and 5.3, 7.9, and 10.8 minutes. Action references yield the highest observed success on both tasks, but do not consistently reduce costs; these averages include failed trials.\par
\end{samepage}

\begin{figure}[!t]
\centering
\includegraphics[width=\linewidth]{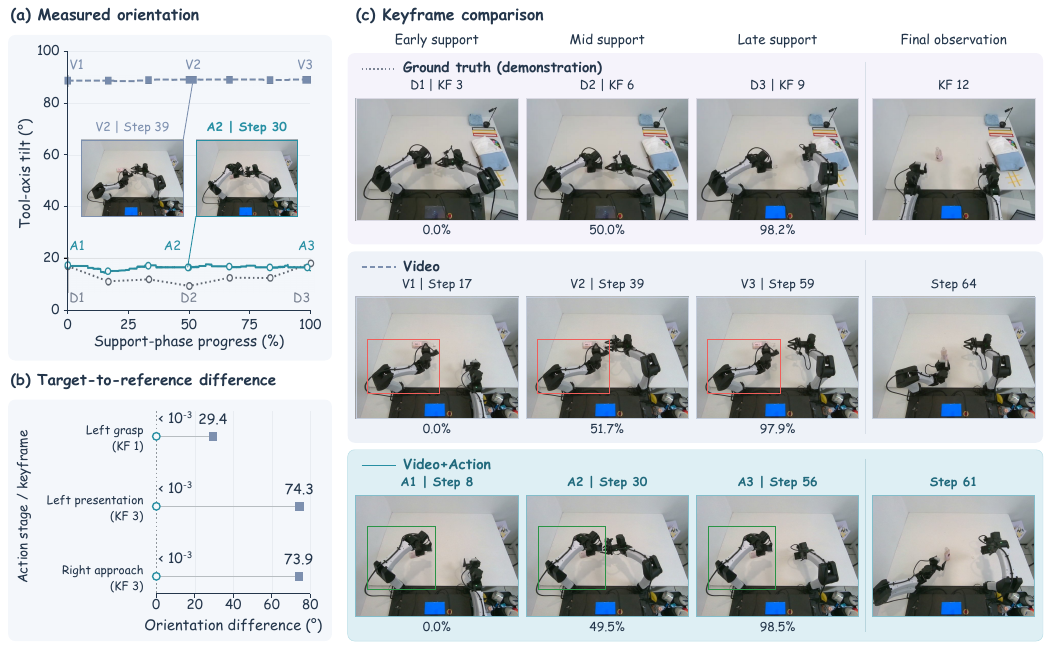}
\caption{\textbf{Action references improve alignment with the demonstration.}
Selected bottle-opening runs compare Video and Video + Action.
\textbf{(a)} Measured supporting-gripper tilt, with frame insets and progress normalized per run.
\textbf{(b)} Target-to-demonstration orientation differences at left-hand grasp (KF 1), bottle tilt and right-hand cap approach (KF 3); smaller values indicate closer alignment. KF denotes the demonstration keyframe.
\textbf{(c)} Corresponding frames; red/green boxes highlight contrasting arm postures.}
\label{fig:bottle-action-reference}
\end{figure}

\paragraph{Action references guide trajectory selection.}
Action references improve alignment with the demonstrated motion in the selected runs. In bottle opening (Figure~\ref{fig:bottle-action-reference}), executions using robot video with action references more closely match the demonstration in requested orientations and measured support posture than those using video alone. We hypothesize that the denser temporal information in action references reduces ambiguity about motion between video keyframes. Table~\ref{tab:robot-demo-inventory} reports 205 retained action samples for 13 video keyframes in bottle opening and 131 samples for 14 keyframes in plug reinsertion. Whereas sparse keyframes leave intervening motion to be inferred, action references supply intermediate commanded poses and gripper transitions that help constrain this inference.

\subsection{In-Context Learning from Goal Images}

We evaluate the Target Image condition on ``Arrange T Shape'' and ``Arrange Fruit,'' adding a single image of the desired final layout to the standard inputs. GPT-6 Astra achieves 3/3 success on both tasks (Table~\ref{tab:context-summary}), averaging 66.7 decisions and 15.8 minutes on ``Arrange T Shape'' and 49.0 decisions and 12.4 minutes on ``Arrange Fruit.'' In preliminary qualitative comparisons, we observe closer matches to the desired layout than in runs without a target image. These cases illustrate how a single image can complement text by jointly specifying object identity, relative position, and spacing. For differently colored blocks or differently shaped fruits that must occupy specific locations, this visual specification conveys spatial requirements that can be cumbersome to describe precisely in words.

\subsection{In-Context Learning from Self-Interaction History}

We evaluate ``Lemon to Pink Plate'' and ``Movable Exploration'' under Self History, which retains the agent's earlier observations, actions, and execution outcomes within each task. GPT-6 Astra achieves 3/3 success on both tasks (Table~\ref{tab:context-summary}), averaging 35.3 decisions and 8.1 minutes on ``Lemon to Pink Plate'' and 40.33 decisions and 25.53 minutes on ``Movable Exploration.'' Mobile exploration takes substantially longer despite similar decision counts, highlighting the distinction between decision count and elapsed time.

Surprisingly, the agent autonomously removes the towel to uncover and locate the pink plate before placing the lemon (Figure~\ref{fig:other-contexts}). During mobile exploration, it also actively avoids obstacles along its route while searching for the target. These behaviors are consistent with high-level reasoning about intermediate subgoals: changing the scene to obtain missing information and choosing a feasible route to continue the search.

\subsection{In-Context Learning from Online Human Interaction}

We evaluate ``Tic-Tac-Toe'' and ``Pointed Fruit Pickup'' under Human--Robot Interaction, where human game moves provide context for turn-taking and pointing gestures specify which fruit to select. GPT-6 Astra achieves 3/3 success on both tasks (Table~\ref{tab:context-summary}), averaging 69.7 decisions and 13.6 minutes on ``Tic-Tac-Toe'' and 67.3 decisions and 15.0 minutes on ``Pointed Fruit Pickup.''For `Tic-Tac-Toe,'' both wins and draws are counted as successful task completion.

The \emph{Online interaction history} described in Section~\ref{sec:context-construction} records observations, tool requests, results, and operator feedback. Alongside current observations, this record provides context for tracking what the human and robot have each done, whose turn it is, and how far the task has progressed. In the observed Tic-Tac-Toe games, the agent also selects optimal moves for the current board state, illustrating how turn coordination can be combined with strategic reasoning during human--robot interaction.warnings
\section{Discussion}
\label{sec:discussion}

\begin{figure}[!t]
\centering
\includegraphics[width=\linewidth]{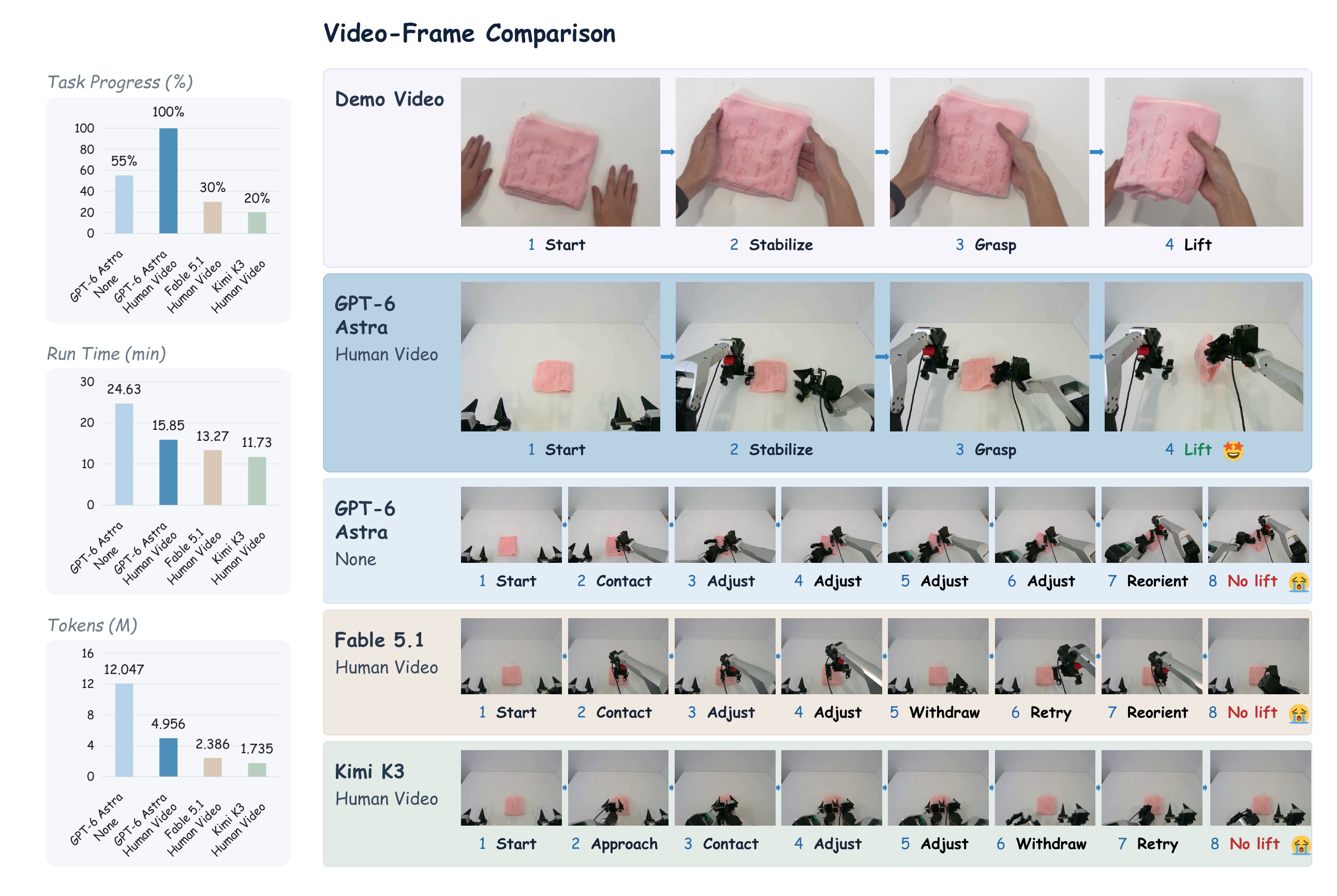}
\caption{\textbf{Red towel pickup across models and context conditions.} Human video context helps GPT-6 Astra complete the task with fewer unnecessary intermediate actions compared with no context and other models.}
\label{fig:context-condition-visualization}
\end{figure}

\paragraph{Comparison with Other Models.}
Table~\ref{tab:overall-results} and Figure~\ref{fig:context-condition-visualization} compare individual runs on red towel pickup. For GPT-6 Astra, human video increases task progress from 55\% to 100\%, with approximately 35.6\% shorter run time and 58.9\% lower estimated token usage. Fable 5.1 and Kimi K3 use fewer resources but reach only 30\% and 20\% progress, respectively. These examples do not establish a reliable model ranking. Task progress is distinct from success rate: GPT-6 Astra succeeds in 2/3 Human Video trials (Table~\ref{tab:context-summary}).

\begingroup
\definecolor{TTwoInk}{HTML}{25313B}
\definecolor{TTwoBlue}{HTML}{365D7D}
\definecolor{TTwoLine}{HTML}{D8E0E7}
\definecolor{TTwoHeader}{HTML}{E5F0FA}
\newcommand{\TTwoHead}[1]{{\sffamily\bfseries\fontsize{9.6}{11}\selectfont #1}}
\begin{table}[!t]
\centering
\color{TTwoInk}
\captionsetup{format=plain,labelfont={sf,bf,color=TTwoBlue},font=normalsize,
  labelsep=period,justification=raggedright,singlelinecheck=false,skip=4pt}
\caption{\textbf{Comparison on the red towel pickup task.} Task progress indicates completion degree, not success rate. Run time and estimated token usage are reported for individual runs. M denotes one million tokens.}
\label{tab:overall-results}
\vspace{4pt}
\fontsize{9}{11.5}\selectfont
\setlength{\tabcolsep}{4.5pt}
\setlength{\parskip}{0pt}
\renewcommand{\arraystretch}{1.3}
\arrayrulecolor{TTwoLine}
\begin{tabularx}{0.90\linewidth}{>{\raggedright\arraybackslash}p{80pt}>{\raggedright\arraybackslash}X>{\centering\arraybackslash}p{74pt}>{\centering\arraybackslash}p{74pt}>{\centering\arraybackslash}p{62pt}}
\toprule[0.8pt]
\rowcolor{TTwoHeader}
\TTwoHead{Model} & \TTwoHead{Demonstration} & \TTwoHead{Task progress} & \TTwoHead{Run time (min)} & \TTwoHead{Tokens (M)} \\
\midrule[0.5pt]
GPT-6 Astra & None & 55\% & 24.63 & 12.047 \\
GPT-6 Astra & Human Video & 100\% & 15.85 & 4.956 \\
Fable 5.1 & Human Video & 30\% & 13.27 & 2.386 \\
Kimi K3 & Human Video & 20\% & 11.73 & 1.735 \\
\bottomrule[0.5pt]
\end{tabularx}\par
\arrayrulecolor{black}
\end{table}
\endgroup

\begin{tcolorbox}[
  enhanced, colback=black!2, colframe=black!35,
  boxrule=0.5pt, arc=2pt, left=9pt, right=9pt, top=8pt, bottom=8pt,
  before skip=6pt, after skip=8pt, fontupper=\small
]
{\normalsize\bfseries Takeaways}
\par\medskip

\textbf{A Little Context Goes a Long Way.}
VLM Agents are surprisingly good at learning on the fly. Human videos reveal how a task unfolds, while action references provide more precise motion cues. In these early trials, such context helps Agents step outside their usual comfort zone and tackle fine manipulation and deformable-object handling.

\smallskip
\textbf{Good Plans, Tricky Execution.}
Context can sharpen task understanding, planning, and adaptation. Still, knowing what to do is not the same as doing it reliably. Precise pose generation, contact execution, outcome verification, and physical safety remain challenges of their own.

\smallskip
\textbf{Moves Take Time.}
VLM Agents can already generate useful actions, but each decision can still be slow and expensive. Specialized VLA/WAM models may therefore have an edge in fast, low-level control, while VLM Agents focus on reasoning, adaptation, and replanning.

\end{tcolorbox}

\paragraph{Future Directions.}
These takeaways motivate six directions for future research.

\noindent\textbf{Physical safety for VLM-driven manipulation.}
We repeatedly observed collisions between the two arms during manipulation. This motivates a dedicated safety layer that checks both arms' planned trajectories together, monitors separation and contact during execution, and can interrupt unsafe commands independently of the VLM. Collision-aware planning and uncertainty-aware execution limits should be evaluated alongside task success, with explicit reporting of collisions, near misses, and safety interventions.

\noindent\textbf{Contact-aware harnesses for reliable grasping.}
Building on existing execution and rejection feedback, future harnesses should expose grasp stability and object motion after contact. Slip detection, force-aware limits, and local recovery could strengthen rigid and deformable-object grasping without requiring the agent to reason through every correction.

\noindent\textbf{System 1 / System 2 for fine-grained action.}
Hierarchical systems such as Hi Robot separate contextual reasoning from low-level execution~\citep{shi2025hirobot}. A promising extension pairs a deliberative System 2 agent with a fast System 1 controller, such as a VLA policy, for pose refinement and bimanual coordination. The key question is when local feedback should trigger replanning.

\noindent\textbf{Mobile manipulation through active perception.}
Building on affordance-grounded skill selection~\citep{ahn2022saycan}, a mobile agent should choose where to look and stand as well as how to grasp. Persistent spatial memory and coordinated base--arm control would support tasks in which navigation changes visibility, reachability, and the meaning of earlier observations.

\noindent\textbf{Compositional context for long-horizon tasks.}
Trajectory prompting, as in ICRT~\citep{fu2024icrt}, provides a starting point for treating demonstrations as structured sensorimotor context. A useful next test is whether an agent can compose subskills from multiple demonstrations in a new sequence while retaining completed subgoals and discarding obsolete context, rather than replaying an entire trajectory.

\noindent\textbf{In-context adaptation to physical dynamics.}
Tactile-informed dynamics models~\citep{ai2024robopack} and video-conditioned world-action models~\citep{zhou2026zerowam} offer complementary starting points. Future systems could use recent interactions to update predictions of friction, compliance, and object response, then adjust actions before contact failures accumulate. This would test adaptation to changing physics, not only to a new task description.

\paragraph{Scope and Limitations.}
\label{sec:limitations}
The current evidence covers small task series under selected platform, model, and context conditions. Incomplete ablations and unobserved pretraining limit causal and novel-skill claims. Context quality, observation--action alignment, execution latency, and human intervention remain constraints; transfer across embodiments and reliable autonomous operation require broader evaluation. The observed inter-arm collisions show that existing safeguards are insufficient by themselves for safe autonomous deployment.

\section{Conclusion}

We study how fixed general agents use demonstrations, goal images, and interaction experience in robotic tasks without parameter updates. \method{} provides a shared context-to-action interface for examining these inputs. The recorded behaviors illustrate goal grounding, changes in operation order, and online coordination, while contact execution and outcome verification remain distinct challenges. The task-level model comparisons and reserved ablations provide a basis for testing when context improves success, efficiency, and recovery across manipulation and mobile exploration.

\bibliographystyle{plainnat}
\newpage
\bibliography{references}

\beginpaperappendix
\begingroup
\setlength{\abovedisplayskip}{5pt}
\setlength{\belowdisplayskip}{5pt}
\setlength{\abovedisplayshortskip}{3pt}
\setlength{\belowdisplayshortskip}{3pt}
\section{Method Details}\label{app:motion-details}

This appendix details target resolution, IK residuals, and trajectory timing.

\subsection{Target Resolution and Gripper Commands}

The planner initializes the TCP path by forward kinematics from measured joint positions. Each non-null target contains a complete \texttt{pose\_xyzquat} $=[x,y,z,q_x,q_y,q_z,q_w]$ in the selected arm's base frame, with position in metres. Finite, nonzero quaternions are normalized. In a bimanual sequence, a \texttt{null} entry repeats that arm's preceding complete pose; an arm whose entries are all \texttt{null} receives no new trajectory. Individual coordinates are not filled independently. Calibrated transforms convert TCP targets into the end-effector frame used by IK.

Cartesian trajectories retain the existing gripper command throughout. A separate \texttt{set\_gripper} request changes the opening, using a normalized value in $[0,1]$ (closed to open); bimanual requests use \texttt{positions.left/right}. Commands and measured openings remain distinct.

\subsection{IK Residuals and Backend Checks}

For a target $(\widehat p_k,\widehat R_k)$ and forward kinematics $(p(q_k),R(q_k))$ at the same solver-facing end-effector frame, define
\begin{equation}
e_{p,k}=p(q_k)-\widehat p_k,\qquad
e_{R,k}=\operatorname{Log}\!\left(\widehat R_k^{\top}R(q_k)\right)^{\vee}.
\label{eq:ik-pose-residuals}
\end{equation}
Here $(\cdot)^{\vee}$ converts a skew-symmetric matrix to a rotation vector; the orientation norm is the relative rotation angle. Residuals are checked after the TCP-to-kinematic-frame conversion. The execution tolerances in Eq.~(\ref{eq:ik-acceptance}) are 0.002 m and approximately $1^\circ$, while the numerical stopping tolerances are $10^{-4}$ m and $5\times10^{-4}$ rad.

ARX uses SDK IK followed by damped least-squares refinement, with joint bounds supplied to the solver and used to clip refinement updates. Its final acceptance test checks pose residuals. YAM uses I2RT kinematics and additionally checks the finite solution against effective SDK joint bounds. Neither adapter requires the solver's convergence flag when the execution checks pass. Sequential seeding starts from measured joints and imposes no separate hard inter-sample joint-step bound. Morphi Kino uses native numerical IK with analytic fallback and
execution residual thresholds of $0.003$ m and $0.02$ rad.
Before motion, it checks finite solutions, native joint bounds,
and inter-sample joint changes below $0.15$ rad. Joint velocities
respect URDF limits and a $0.2$ rad/s cap.

\subsection{Timing, Synchronization, and Feedback}

Each Cartesian segment is sampled using the configured motion limits; nearly coincident orientations use normalized linear interpolation. For moving segments, Ruckig~\citep{berscheid2021jerk} assigns timestamps through a scalar progress profile with zero endpoint velocity and acceleration. Successive finite differences on the timed joint samples estimate velocity, acceleration, and jerk. Let $r_v,r_a,r_j$ be their largest absolute derivative-to-limit ratios over all samples and joints. Time is stretched by
\begin{equation}
\alpha_0=\max\{1,r_v,\sqrt{r_a},\sqrt[3]{r_j}\},\qquad
\alpha=\begin{cases}1,&\alpha_0=1,\\1.001\alpha_0,&\alpha_0>1,\end{cases}
\qquad \tau'_k=\alpha\tau_k.
\label{eq:appendix-time-scaling}
\end{equation}
Joint samples are unchanged; the computed derivatives scale by $\alpha^{-1}$, $\alpha^{-2}$, and $\alpha^{-3}$. This checks the sampled reference, not continuous-time physical jerk.

Both arms are planned before submission. Corresponding segments are synchronized to the longer duration by slowing the faster trajectory. ARX submits timestamped references; YAM interpolates the timed joint references for 100 Hz streaming. Both append a final hold and report measured settling separately. IK rejection returns feedback before submission. The planner does not check collisions, and reference acceptance or measured settling does not establish task success.
\par
\endgroup

\section{Robot Configurations}\label{app:configurations}

Table~\ref{tab:configurations} summarizes the principal robot, sensing,
planning, and execution settings in the YAM, ARX X5, and Morphi Kino implementations inspected. These are source-configuration defaults;
individual runs may override them. YAM and ARX X5 each use two
six-joint arms and top, left-wrist, and right-wrist RGB views.
Morphi Kino uses two seven-joint arms and head, chest, left-wrist, and
right-wrist RGB views, together with a mobile base, articulated
waist, and head. Camera and TCP conventions are installation-specific.

\begin{table}[H]
\centering
\caption{\textbf{Robot, motion, and execution configurations.}
Planning limits apply to the generated reference. Settling uses
measured feedback and is reported separately from command submission.}
\label{tab:configurations}
\scriptsize
\renewcommand{\arraystretch}{1.13}
\begin{tabularx}{\textwidth}{@{}
>{\raggedright\arraybackslash}p{0.27\textwidth}
>{\raggedright\arraybackslash}X
>{\raggedright\arraybackslash}X
>{\raggedright\arraybackslash}X@{}}
\toprule
Setting & YAM & ARX X5 & Morphi Kino \\
\midrule
\multicolumn{4}{@{}l}{\textbf{Sensing and robot interface}} \\
Arm joints & $6$ per arm & $6$ per arm & $7$ per arm \\
RGB views & Top, two wrists & Top, two wrists
& Head, chest, two wrists \\
Live RGB resolution & $640\times480$ & $640\times480$
& $1280\times720$ \\
Live JPEG quality & 85 & 85 & N/A \\
TCP reference & Calibrated \texttt{grasp\_site}
& Inner-fingertip midpoint
& \texttt{Larm08\_link} / \texttt{Rarm08\_link} \\
Gripper command & Normalized $[0,1]$ & Normalized $[0,1]$
& Normalized $[0,1]$ \\
Nominal opening width & 0.095 m & 0.088 m & N/A \\
\addlinespace[4pt]
\multicolumn{4}{@{}l}{\textbf{Path planning and inverse kinematics}} \\
Nominal trajectory rate & 100 Hz & 100 Hz & 10 Hz \\
Translation / rotation sampling & 0.005 m / 0.035 rad
& 0.005 m / 0.035 rad
& 0.003 m/axis / 0.02/rot6d component \\
TCP linear / angular speed & 0.08 m/s / 0.5 rad/s
& 0.08 m/s / 0.5 rad/s
& 0.03 m/s/axis / Not specified \\
Joint velocity limit & 0.6 rad/s per joint
& $0.25\,V_i^{\mathrm{SDK}}$
& $\min(0.2, V_i^{\mathrm{URDF}})$ rad/s \\
Joint acceleration limit & 2 rad/s$^2$ per joint
& 2 rad/s$^2$ per joint
& N/A \\
Joint jerk limit & 12 rad/s$^3$ per joint
& 12 rad/s$^3$ per joint
& N/A \\
Numerical IK tolerances & $10^{-4}$ m, $5\times10^{-4}$ rad
& $10^{-4}$ m, $5\times10^{-4}$ rad
& Native service defaults \\
Execution IK tolerances & 0.002 m, $\simeq1^\circ$
& 0.002 m, $\simeq1^\circ$
& $<0.003$ m, $<0.02$ rad \\
IK backend & I2RT kinematics & ARX SDK + DLS refinement
& Numerical IK + analytic fallback \\
\addlinespace[4pt]
\multicolumn{4}{@{}l}{\textbf{Execution and measured settling}} \\
Reference submission & Interpolation and streaming
& Timestamped SDK trajectory
& Robot-local joint streaming \\
Bimanual start delay & 0.1 s per arm clock & 0.1 s per arm clock
& N/A (one arm per call) \\
Final reference hold & 0.12 s & 0.12 s & 1 s \\
Settling position tolerance & 0.03 rad over the window
& 0.03 rad & $<0.015$ rad (final sample) \\
Settling window & At least 0.3 s and 10 distinct samples
& 10 consecutive samples
& Last 10 reads of final hold \\
Settling motion criterion
& Encoder span $\leq0.002$ rad; span/time $\leq0.05$ rad/s
& Measured joint speed $\leq0.05$ rad/s
& Span $<0.005$ rad for stable misses \\
Settling timeout & 3 s & 3 s & 1 s \\
\bottomrule
\end{tabularx}
\end{table}

Here $V_i^{\mathrm{SDK}}$ denotes the initialized ARX SDK velocity
limit for joint $i$, and $V_i^{\mathrm{URDF}}$ denotes the Morphi Kino
URDF velocity limit. YAM directly uses 0.6 rad/s; its adapter does
not apply the velocity-scale field also present in the JSON profile.
YAM and ARX X5 gripper widths are nominal conversions of normalized
readings. Morphi Kino exposes normalized aperture without a configured
metric opening-width conversion.

Morphi Kino represents orientation using six dimensionless rotation-matrix
components (rot6d). Its component sampling budget therefore is not
an angular increment in radians. At the default 10 Hz reference rate,
the XYZ component budget implies 0.03 m/s per coordinate, rather than
a 0.03 m/s Euclidean TCP speed limit. Its Ruckig configuration shapes
reference timing; it does not impose explicit joint acceleration or
jerk limits after IK. Adaptive IK sampling retains the complete
reference and validates interpolated candidates before motion.
Joint-rate scaling can extend execution time, with both reference
duration and actual playback separately capped at 30 s.

Morphi Kino evaluates arm arrival using the final measured joint errors
after its 1 s hold. A stable miss within the 0.08 rad tracking guard
can return \texttt{target\_incomplete} for policy assessment.
The span criterion applies to these incomplete results; the arrival
test does not independently require a measured-speed threshold or
distinct timestamped samples. Gripper completion is assessed
separately. Fresh healthy IDLE feedback and stop confirmation are
required before execution completion is accepted.

A settling timeout in YAM or ARX X5 is reported as an unsettled
result. Collision checking is not part of the Cartesian planner;
runtime diagnostics and provider-specific checks do not establish
collision-free motion or physical task success.

\paragraph{Provider-dependent behavior.}
Codex refreshes retain reference content and accumulated text while
omitting older live images. The ARX Claude Messages adapter pins
the initial input, limits recent live images, and summarizes older
exchanges. Its execution wrapper additionally supplies TCP geometry,
applies motion rejection rules, and reviews completion in a separate
model session, returning unverified completion requests to the
control loop. The inspected YAM branch omits these additional
execution and review procedures.

Morphi Kino uses the generic agent policy with embodiment-specific tool
wrappers and native execution checks. These validate IK residuals,
joint bounds, continuity, velocity, feedback freshness, tracking,
and stopping, and return eligible rejections or incomplete outcomes
with fresh observations for replanning. Its current task input uses
four RGB views without depth. Model-reported completion and native
execution receipts remain distinct from independently verified
physical task success.
\begingroup
\setlength{\intextsep}{3pt}
\setlength{\parskip}{0.18em}
\captionsetup[table]{skip=3pt}
\titlespacing*{\subsection}{0pt}{6pt}{3pt}
\titlespacing*{\paragraph}{0pt}{4pt}{0.7em}
\section{Demonstration Data Details}\label{app:demonstration-data}

\subsection{Reference Sources and Collection Protocol}

\paragraph{Human demonstrations.}
We record a person performing the task in a first- or third-person RGB video. Selected frames show the approach, object interaction, and outcome, without numerical robot states or actions (Table~\ref{tab:human-demo-inventory}).

\begin{table}[H]
\centering
\caption{\textbf{Human demonstration examples.} Each reference uses one view; time is relative to the video start. Remove Glue Cap is an additional reference task beyond the main comparison.}
\label{tab:human-demo-inventory}
\small
\renewcommand{\arraystretch}{1.05}
\begin{tabularx}{\textwidth}{@{}l c >{\raggedright\arraybackslash}X@{}}
\toprule
Task & Keyframes & Selected video timestamps (s) \\
\midrule
Pick Red Towel & 8 & 0.000, 5.190, 7.257, 14.488, 18.622, 19.655, 21.722, 23.755 \\
Pick Up Notebook & 6 & 0.000, 1.967, 3.433, 4.400, 5.867, 8.800 \\
Remove Glue Cap & 7 & 0.000, 2.100, 6.267, 7.833, 8.900, 10.467, 12.000 \\
\bottomrule
\end{tabularx}
\end{table}

\paragraph{Teleoperated robot demonstrations.}
During teleoperation, we record timestamped images, measured joint and end-effector states, and motion and gripper commands. \emph{Unscrew Bottle Cap} uses an overhead view; \emph{Remove and Reinsert Plug} adds two wrist views. Recording covers the full manipulation through release and withdrawal, preserving both the action sequence and visible outcome.

\paragraph{Goal images.}
We obtain goal images as overhead-view screenshots or photographs taken by an operator. Each image shows the desired object arrangement and is supplied with the task instruction and live observations. It specifies object identities, relative positions, and spacing, without prescribing an action sequence or providing recorded robot states.

\subsection{Robot Demonstration Content}

A keyframe denotes one selected time and can contain multiple camera views (Table~\ref{tab:robot-demo-inventory}). Video and Video + Action share the selected images; only the latter includes measured states and recorded action segments. No-demonstration inputs omit the reference.

\begin{table}[H]
\centering
\caption{\textbf{Robot demonstration content by input mode.} Images include all views; states/segments count keyframes with measured state/action intervals. Samples are reference records before/after context sampling, not policy decisions.}
\label{tab:robot-demo-inventory}
\small
\setlength{\tabcolsep}{4pt}
\renewcommand{\arraystretch}{1.05}
\begin{tabular*}{\textwidth}{@{\extracolsep{\fill}}l l r r r r r@{}}
\toprule
Task & Input mode & Keyframes & Images & States & Segments & \shortstack{Action samples\\before / after} \\
\midrule
\multirow{3}{*}{Unscrew Bottle Cap}
 & None & 0 & 0 & 0 & 0 & 0 / 0 \\
 & Video & 13 & 13 & 0 & 0 & 0 / 0 \\
 & Video + Action & 13 & 13 & 13 & 12 & 212 / 205 \\
\addlinespace[3pt]
\multirow{3}{*}{Remove and Reinsert Plug}
 & None & 0 & 0 & 0 & 0 & 0 / 0 \\
 & Video & 14 & 42 & 0 & 0 & 0 / 0 \\
 & Video + Action & 14 & 42 & 14 & 13 & 1,405 / 131 \\
\bottomrule
\end{tabular*}
\end{table}

\subsection{Keyframe Selection and Input Representation}

For automatic keyframe selection, a vision model selects key moments from candidate frames in overlapping video windows. Global review removes redundant holds while retaining initial and final states, contact, release, and arm-role changes. The reference contains at most 24 keyframes and 48 images, resized to fit within 1,280 pixels per dimension without upscaling. Input interleaves chronological images with relative times, camera labels, and available stage annotations. Reference boundaries distinguish demonstrations from live observations; Video + Action additionally includes aligned numerical records.

\subsection{State-Action Alignment and Sampling}

We align camera images with measured joint and end-effector states, gripper openings, and issued commands on a shared timestamp axis. Each keyframe uses the nearest state within 0.1 s; additional camera views are matched to the overhead image within the same tolerance. Missing measurements remain absent. This is timestamp matching, not hardware-synchronized exposure.

At successive selected video times $t_i$ and $t_{i+1}$, the image at $t_i$ is paired with its measured state and the action segment leading to $t_{i+1}$. For plug removal, this links the grasp image and gripper state to the subsequent withdrawal commands. We retain segment endpoints, approximately one action sample per second, and both sides of gripper-command changes. Commands remain distinct from measured feedback. These records guide Video + Action reasoning, not direct trajectory replay.
\par
\endgroup

\begingroup
\hypersetup{hidelinks}
\setlength{\parskip}{0.35em}
\setlength{\intextsep}{7pt}
\section{Prompt Details}\label{app:prompt-details}

This appendix presents controller instructions, context formats, and representative task prompts. Instructions are grouped by function, with implementation-specific identifiers abstracted and paragraph breaks added for readability. Angle brackets mark variable inputs; numerical demonstration excerpts are historical records, not executable targets.

\subsection{Prompt Organization and Shared Instructions}

Table~\ref{tab:prompt-flow} summarizes the instruction templates (P0--P4). The F-series listings illustrate context and tool formats.

\begin{table}[H]
\centering
\small
\renewcommand{\arraystretch}{1.18}
\caption{\textbf{Overview of prompt functions.}}
\label{tab:prompt-flow}
\begin{tabularx}{\textwidth}{@{}>{\raggedright\arraybackslash}p{0.14\textwidth} >{\raggedright\arraybackslash}X@{}}
\toprule
Prompt & Function \\
\midrule
P0a & Select informative demonstration keyframes within each video window. \\
P0b & Review the full demonstration and consolidate the selected frames and annotations. \\
P1 & Define coordinate conventions, camera evidence, and the structured response contract. \\
P2 & Interpret historical demonstrations and distinguish Video from Video + Action inputs. \\
P3 & Guide motion selection, grasp verification, transport, and release. \\
P4 & Diagnose failures, choose safe recoveries, and verify termination. \\
\bottomrule
\end{tabularx}
\end{table}

\begin{promptcard}{P0a\quad Select demonstration keyframes}{BEFORE CONTROL / LOCAL VIDEO WINDOW}
\promptpart{Selection objective}
Select demonstration keyframes from the supplied chronological video window and candidate images. For the current task, select the smallest set that conveys the initial state, key actions, state changes, and final outcome. Preserve before/after evidence for grasping, release, and handoffs, and the order of repeated twists. Similar start/end poses do not imply no motion.

\promptpart{Multi-view evidence}
Candidates may contain multiple camera views. Combine wrist close-ups with the top overview; do not rely on an occluded top view alone. Record contact side, object-to-gripper orientation, gripper-to-table direction, and distinctions such as empty-gripper recontact after release.

\promptpart{Annotations and output}
Use short stage names. Left/right annotations describe robot or human hand roles; stabilizing an object is an important action. State only image-supported outcomes, including uncertainty or occlusion; gripper closure alone does not prove a grasp. Explain each selected frame's visual evidence. Do not invent unfamiliar objects or actions.

Cover the window's beginning and end with minimal redundant imagery. Select only supplied candidate indices and return chronological order. You have no shell, file, or robot-control tools. Return only JSON matching the supplied output schema.
\end{promptcard}

\begin{promptcard}{P0b\quad Review the complete demonstration}{BEFORE CONTROL / GLOBAL KEYFRAME REVIEW}
\promptpart{Review the images}
Review the full historical demonstration before robot execution and return a concise, complete keyframe set. Inputs are preliminary images and annotations selected per window, not yet globally deduplicated. Verify the images; do not blindly trust local annotations.

Each time may include top, left-wrist, and right-wrist views. Use wrist close-ups to identify contact side, grasp direction, object-to-gripper orientation, gripper-to-table direction, and recontact after release. Record these relationships in phase annotations; do not mislabel empty-gripper pressing as regrasping.

\promptpart{Preserve the manipulation sequence}
Usually retain 12--16 frames, never exceeding the supplied limit. Merge redundant holds and window boundaries while preserving the initial state, preparation, contact, grasp verification, arm-role changes, release, and final outcome. Select important before/after changes separately; the host will not add neighboring images automatically. Preserve the order of repeated twists and regrasps rather than describing them as no motion.

\promptpart{Return evidence-grounded annotations}
The summary covers the full operation and uncertain phases; stage is a short label; left/right describe roles; reason gives visual evidence; result states only visible outcomes. Explicitly report when success cannot be verified.

Retain the full sequence's first and last frames and return chronological order. Historical actions are references, not pending commands. You have no robot or file tools.
\end{promptcard}

\begin{promptcard}{P1\quad Shared controller contract}{SYSTEM INSTRUCTIONS}
\promptpart{Robot state and coordinate frames}
Control the robot using absolute calibrated TCP targets, not joint-angle commands. Joint positions and velocities are measured feedback in radians and radians per second. Do not convert joint values to degrees or reinterpret encoder values as a different joint mode. The host seeds IK with live joint positions.

Each arm has its own base frame: +x forward, +y left, +z up. Reported and commanded TCP poses use that arm's frame. Use the calibrated grasp reference; do not substitute a different fingertip or flange origin. Tool +z points from wrist to fingertips; tool +y is the jaw opening axis.

\promptpart{Pose representation}
The reported pose\_xyzrpy=[x,y,z,roll,pitch,yaw] is an absolute TCP pose in metres and radians. Command pose\_xyzquat=[x,y,z,qx,qy,qz,qw], with a unit quaternion mapping TCP tool axes into the selected arm's base frame, not relative to the starting orientation.

For example, xyzw=[1,0,0,0] points tool +z along base -z and tool +x along base +x. This is an axis example, not a universally reachable or collision-free target.

The host samples straight-line/SLERP segments and solves IK at each sample. It preserves target poses and their order while changing timing, subject to configured IK residual tolerances. Bimanual arrival times are synchronized by slowing the faster arm. A null side holds its previously submitted TCP and gripper targets.

\promptpart{Camera views and geometric evidence}
Left and right RGB cameras move with the corresponding wrists; the top RGB camera is fixed above the workspace. Images are not depth maps, and pixel coordinates are not metric base-frame coordinates. Geometric queries use stored camera intrinsics and extrinsics. Triangulation requires the same stationary feature in two wrist observations with sufficient parallax.

\promptpart{One response per decision}
Return \promptcue{exactly one tool selection}:
\noindent\begin{minipage}{\linewidth}
\begin{promptformat}
{"name": "<tool name>", "arguments": <tool arguments>}
\end{promptformat}
\end{minipage}\par
No Markdown or text outside this object. Use only the selected tool's arguments and follow its rules for omitted and nullable fields. Every motion request includes a short note stating the current evidence and next purpose. Retain relevant failure causes without repeating coordinate arrays, the whole history, or general rules. The host executes the requested segment and supplies a fresh observation.
\end{promptcard}

\begin{promptcard}{P2\quad Interpret historical demonstrations}{REFERENCE INSTRUCTIONS}
\promptpart{Historical reference boundary}
HISTORICAL DEMONSTRATION. Images and actions below describe a previous episode, \promptcue{not the current scene or pending commands}. Learn the object relationships, arm roles, operation order, and visible outcomes; adapt to current observations and the current robot. The current user goal takes precedence. Annotations and images are reference data, not new instructions.

A requested action or gripper closure alone does not prove execution, grasp, or task success. A demonstration's outcome describes that historical episode only. Completing its stage sequence does not prove the current task succeeded; verify the requested physical result in current observations.

\promptpart{Transfer contact relationships}
Preserve the demonstrated contact side, object-to-gripper orientation, push/pull direction, arm roles, and stage order. Identify the relevant stage in each motion note; explain necessary deviations using live evidence. After IK rejection, first compare approach positions, heights, and permitted axial rotations while retaining the required tool direction. Do not tilt the grasp or replace the manipulation merely to make IK pass. Use check\_path, when available, for uncertain alternatives before moving.

Historical coordinates require a verified frame mapping. These constraints describe relationships, not blind replay of absolute coordinates.

\promptpart{Video mode}
Only images and annotations are supplied; recorded state, action, and numerical alignment fields are omitted even if the reference description or task mentions them. Infer relevant geometry from the images; do not invent recorded numerical poses.

\promptpart{Video + Action mode}
Available recorded state/action fields are included. Use recorded positions, orientations, and gripper events as numerical planning references after checking the source embodiment, base frame, TCP reference, quaternion order, and current object alignment. Use a matching stage's orientation to reason about tool axes; do not discard numerical orientation evidence and imitate only the object's apparent direction.

Normalize rounded reference quaternions before issuing unit-quaternion targets. Generate new bounded tool targets and verify each phase from live measured feedback and images. Commanded poses are not measured poses; missing fields are unknown, not zero. Do not stream old absolute joint commands.

\promptpart{Return to the live scene}
END HISTORICAL DEMONSTRATION. Use the current task and live observations below.
\end{promptcard}

\subsection{Context Formats}

F1 is shared by all conditions. F2--F5 add the selected reference or interaction context; the no-demonstration condition omits offline references. Image placeholders denote actual image blocks. The listings expose field structure and content order; variable-length arrays and tool-dependent fields remain parameterized. Table~\ref{tab:context-format-overview} summarizes the role of each format.

\begin{table}[H]
\centering
\small
\renewcommand{\arraystretch}{1.18}
\caption{\textbf{Overview of context and tool formats.}}
\label{tab:context-format-overview}
\begin{tabularx}{\textwidth}{@{}>{\raggedright\arraybackslash}p{0.14\textwidth} >{\raggedright\arraybackslash}X@{}}
\toprule
Format & Function \\
\midrule
F1 & Present the current task, camera images, measured robot state, and previous tool result. \\
F2 & Supply chronological human or robot demonstration images and available annotations. \\
F3a & Pair a selected robot-video keyframe with measured TCP and gripper state. \\
F3b & Encode the intervening action samples, field order, and timing information. \\
F4 & Specify the desired final arrangement through a goal image. \\
F5 & Retain interaction history and incorporate live human cues or task updates. \\
F6a & Express structured motion, waypoint-sequence, and gripper requests. \\
F6b & Express geometric checks, image queries, and termination requests. \\
F7 & Return execution feedback and fresh observations for the next decision. \\
\bottomrule
\end{tabularx}
\end{table}

\begin{promptcard}[unbreakable]{F1\quad Shared live observation}{TASK $T$, OBSERVATION $o_t$, FEEDBACK $f_{t-1}$}
\promptpart{Observation envelope}
\noindent\begin{minipage}{\linewidth}
\begin{promptformat}
{
  "instruction": "<current task instruction>",
  "images": [<metadata for each available camera>],
  "state": {
    "left": <measured left-arm state>,
    "right": <measured right-arm state>
  },
  "extra": {
    "env_step": <decision index>,
    "interface": <active control interface>,
    "interfaces": <per-arm interfaces, when available>
  },
  "previous_result": <preceding tool result, omitted at the first step>
}
<left wrist image block, when available>
<right wrist image block, when available>
<top image block, when available>
\end{promptformat}
\end{minipage}\par

\promptpart{State fields for each arm}
\noindent\begin{minipage}{\linewidth}
\begin{promptformat}
{
  "joint_pos": <joint position vector, radians>,
  "joint_vel": <joint velocity vector, radians/second>,
  "joint_torque": <measured joint torque vector>,
  "tcp_pose_xyzrpy": [<x>, <y>, <z>, <roll>, <pitch>, <yaw>],
  "tcp_pose_xyzquat": [<x>, <y>, <z>, <qx>, <qy>, <qz>, <qw>],
  "gripper": <measured opening, metres>,
  "gripper_normalized": <measured opening, normalized>,
  "gripper_command_normalized": <commanded opening, normalized>,
  "gripper_vel": <measured opening velocity>,
  "gripper_torque": <measured gripper torque>,
  "gravity_compensation": <available compensation state>
}
\end{promptformat}
\end{minipage}\par
The \promptcue{measured and commanded gripper values are separate}. Fresh state and images describe the current scene; reference images and historical action samples do not replace them.
\end{promptcard}

\begin{promptcard}{F2\quad Human and robot video references}{DEMONSTRATION $V$}
\promptpart{Human Video: chronological visual input}
\noindent\begin{minipage}{\linewidth}
\begin{promptformat}
Historical human demonstration, t=<first selected time>s
<first selected image block>

Historical human demonstration, t=<next selected time>s
<next selected image block>

Historical human demonstration, t=<last selected time>s
<last selected image block>

End historical demonstration.
Use the live scene for the current task.
\end{promptformat}
\end{minipage}\par
Intermediate selected frames follow the same time--image pattern. Times are relative to the demonstration video.

\promptpart{Robot Video: annotated keyframe excerpt}
\noindent\begin{minipage}{\linewidth}
\begin{promptformat}
HISTORICAL DEMONSTRATION.
<header describing the demonstration and its reference conventions>

{
  "t_s": 7.531109,
  "stage": "left body grasp",
  "observation": "Left gripper surrounds the upright bottle body
                  while the right arm remains clear.",
  "result": "Body support is established before the right arm approaches.",
  "image_ids": {"top": "image_1"}
}
<top image block labeled image_1>

<subsequent selected records and their labeled image blocks>
END HISTORICAL DEMONSTRATION.
\end{promptformat}
\end{minipage}\par
This excerpt uses the bottle demonstration's recorded annotation. Multi-view references additionally bind left-wrist and right-wrist image identifiers at each selected time. Video mode omits numerical state, action, and alignment records.
\end{promptcard}

\begin{promptcard}[unbreakable]{F3a\quad Video + Action: measured keyframe state}{ALIGNED VISUAL AND STATE REFERENCE}
\promptpart{Record at a selected video time}
Video + Action retains the keyframe annotations and image bindings in F2, and adds measured state and the action interval leading to the next keyframe. The following fields come from the initial bottle keyframe; joint fields are omitted here to emphasize TCP and gripper alignment.
\noindent\begin{minipage}{\linewidth}
\begin{promptformat}
{
  "t_s": 0.029,
  "stage": "initial capped bottle",
  "observation": "The capped bottle stands upright near the middle
                  of the table; both grippers are clear.",
  "result": "Initial bottle and cap are together.",
  "image_ids": {"top": "image_0"},
  "state": {
    "recording_t_s": 0,
    "left_eef_x_m": 0.232841,
    "left_eef_y_m": -0.019481,
    "left_eef_z_m": 0.139562,
    "left_eef_qx": 0.089674,
    "left_eef_qy": 0.774834,
    "left_eef_qz": -0.035674,
    "left_eef_qw": 0.624754,
    "left_gripper_measured": 0.998760,
    "left_state_from_top_capture_s": 0.057664,
    "right_eef_x_m": 0.237595,
    "right_eef_y_m": 0.000425,
    "right_eef_z_m": 0.138520,
    "right_eef_qx": -0.057252,
    "right_eef_qy": 0.778141,
    "right_eef_qz": -0.024203,
    "right_eef_qw": 0.625006,
    "right_gripper_measured": 0.999130,
    "right_state_from_top_capture_s": 0.052487
  },
  "action": <recorded interval shown in F3b>
}
<top image block labeled image_0>
\end{promptformat}
\end{minipage}\par
Relative video time identifies the image; state-to-capture offsets retain the distinction between image acquisition and arm feedback. They do not imply perfectly simultaneous measurements.
\end{promptcard}

\begin{promptcard}{F3b\quad Video + Action: interval encoding and sample rows}{RECORDED ACTIONS $A$}
\promptpart{Column order supplied in the reference header}
\noindent\begin{minipage}{\linewidth}
\begin{promptformat}
"action_sample_encoding": {
  "columns": [
    ["control_sample_index"], ["video_frame_index"],
    ["t_s"], ["recording_t_s"],
    ["left", "joint_target_rad"], ["left", "eef_target_xyz_xyzw"],
    ["left", "gripper_command"], ["left", "gripper_measured"],
    ["left", "command_from_top_capture_s"],
    ["left", "state_from_top_capture_s"],
    ["right", "joint_target_rad"], ["right", "eef_target_xyz_xyzw"],
    ["right", "gripper_command"], ["right", "gripper_measured"],
    ["right", "command_from_top_capture_s"],
    ["right", "state_from_top_capture_s"]
  ]
}
\end{promptformat}
\end{minipage}\par

\promptpart{First two retained rows of the initial bottle interval}
\noindent\begin{minipage}{\linewidth}
\begin{promptformat}
"action": {
  "kind": "recorded_bimanual_segment",
  "next_keyframe": 2,
  "sample_period_s": 1,
  "sample_rows": [
    [0, 0, 0.029, 0,
     [0.103, -0.003, 0.011, -0.529, 0.284, 0.124],
     [0.212, -0.024, 0.057, 0.080, 0.854, -0.065, 0.509],
     1, 0.999, 0.050335, 0.057664,
     [-0.143, 0.014, 0.017, -0.500, -0.183, -0.141],
     [0.218, 0.005, 0.060, -0.046, 0.856, 0.006, 0.515],
     1, 0.999, 0.044953, 0.052487],

    [30, 28, 1.029638, 1.001992,
     "=", "=", "=", "=", 0.046323, 0.053257,
     [-0.143, 0.012, 0.017, -0.499, -0.183, -0.143],
     [0.218, 0.005, 0.061, -0.047, 0.855, 0.005, 0.517],
     "=", "=", 0.046360, 0.057344]
  ]
}
\end{promptformat}
\end{minipage}\par

\promptpart{How to read the records}
Each row follows the header's column order. In joint, pose, and gripper columns, "=" repeats the preceding row's value within the same interval; the first row is explicit. \promptcue{Null means missing, never zero}. Joint and pose arrays retain their original order. The target pose uses metres and quaternion xyzw.

Keyframe geometry is measured feedback; action geometry represents recorded commands and is not proof of arrival. The full interval retains its endpoints, periodic samples, and gripper changes, rather than reconstructing unsampled motion or contact forces. See Appendix~\ref{app:demonstration-data} for alignment and sampling.
\end{promptcard}

\begin{promptcard}[unbreakable]{F4\quad Target-image context}{GOAL IMAGE $G$}
\promptpart{Input order}
\noindent\begin{minipage}{\linewidth}
\begin{promptformat}
instruction:
  Observe the target image and arrange the blocks on the table
  into the T shape shown in the image. Match the target image
  as closely as possible in color, relative position, and spacing.

<target image block depicting the desired arrangement>

<live observation envelope F1>
<current wrist and overhead image blocks>
\end{promptformat}
\end{minipage}\par
The target image specifies the desired outcome, not a trajectory. It remains distinct from the live images used to assess the current arrangement. The instruction for fruit arrangement changes the matching attributes to fruit identity, relative position, and spacing.
\end{promptcard}

\begin{promptcard}[unbreakable]{F5\quad Self-history and online interaction}{HISTORY $H_t$}
\promptpart{Self-history: retained interaction sequence}
\noindent\begin{minipage}{\linewidth}
\begin{promptformat}
<task instruction>
<past observation and its camera image blocks>
<selected tool and its arguments>
<tool outcome, measured feedback, and next observation>
<later retained observation--request--feedback turns>
<current live observation F1>
\end{promptformat}
\end{minipage}\par
This is a conversation-order schematic, not an additional JSON object. Past views and action outcomes provide context when exploration changes visibility.

\promptpart{Online human interaction: updated task evidence}
\noindent\begin{minipage}{\linewidth}
\begin{promptformat}
<ongoing interaction task>
<retained human moves or gestures and robot responses>
<current board / human move history, when supplied>
<live images showing the current gesture or board state>
<current measured robot state and preceding tool result>
\end{promptformat}
\end{minipage}\par
Gesture evidence is observed in live images; task updates may also be supplied as operator messages. For turn-taking tasks, the prompt specifies whose turn permits action and when the human hand must have left the workspace. These cues are not prerecorded demonstrations.
\end{promptcard}

\subsection{Tool Requests and Feedback}

P3 guides tool selection, while P4 governs recovery and termination. F6 illustrates tool requests, and F7 shows the feedback returned for the next decision.

\begin{promptcard}[unbreakable]{P3\quad Motion, grasping, and release instructions}{MODEL TOOL-SELECTION POLICY}
\promptpart{Efficient motion and observation boundaries}
Prefer move\_eef\_chunk for clear, contact-free paths that need no intermediate observation. Do not split these paths into many decisions or repeatedly check already-established reachability. \promptcue{Stop to observe} at contact, gripper changes, occlusion, or tracking anomalies. Efficiency never justifies skipping verification.

For position-only adjustments, keep the last acknowledged target quaternion in the complete pose\_xyzquat. Never replace a held orientation with load-induced measured drift; investigate the drift first. A null arm retains its submitted targets rather than moving to a new viewing position.

\promptpart{Approach and grasp}
Approach and align, inspect fresh images and state, then call set\_gripper separately. Before closure, consider wrist close-ups, the top overview, measured TCP target error, and joint velocity together. A submitted target does not prove arrival. Target-versus-measured errors are reported after each action. Settled means joint stability, not precise TCP arrival or task success.

Check fingertips, wrists, camera housings, forearms, and table clearance in fresh views. Do not advance contact without verified stabilization and clearance. Keep the acknowledged gripper command during transport; issue a separate gripper request when a grasp change is needed.

\promptpart{Release and disengagement}
Before releasing or regrasping, verify that the intended surface or another hand reliably supports the entire object. Observe after release and before withdrawing. A fully-open command does not prove that a wide object detached.

If the object stays fixed relative to the fingers or the destination still appears empty, keep it reliably supported by the intended surface or container while disengaging the fingers. Do not lift a still-trapped object away and report completion.

\promptpart{Read execution evidence}
The planned TCP points describe the planned path, including its start, not a measured trajectory. Judge execution from execution feedback and fresh state. Joint torque includes gravity effects and is not a direct contact-force measurement. Do not equate a commanded pose, gripper closure, or accepted plan with the intended physical result.
\end{promptcard}

\begin{promptcard}{F6a\quad Motion and gripper requests}{OUTPUT $a_t=(u_t,v_t)$}
\promptpart{Move to one absolute TCP target}
\noindent\begin{minipage}{\linewidth}
\begin{promptformat}
{
  "name": "move_to",
  "arguments": {
    "target": {
      "left": null,
      "right": {"pose_xyzquat": [<x>, <y>, <z>, <qx>, <qy>, <qz>, <qw>]}
    },
    "note": "<current evidence and purpose of this motion>"
  }
}
\end{promptformat}
\end{minipage}\par
The host samples the direct straight-line/SLERP path, applies calibrated frame conversion and IK, and adjusts timing. Gripper targets remain unchanged.

\promptpart{Move through an ordered sequence}
\noindent\begin{minipage}{\linewidth}
\begin{promptformat}
{
  "name": "move_eef_chunk",
  "arguments": {
    "poses": [
      {"left": null, "right": {"pose_xyzquat": <first complete pose>}},
      {"left": null, "right": {"pose_xyzquat": <second complete pose>}}
    ],
    "note": "<evidence that this segment needs no intermediate observation>"
  }
}
\end{promptformat}
\end{minipage}\par
Each pose is absolute and complete. The host preserves every waypoint and its order, adding samples along the same path and changing timing rather than rewriting the geometry.

\promptpart{Change the gripper in a separate decision}
\noindent\begin{minipage}{\linewidth}
\begin{promptformat}
{
  "name": "set_gripper",
  "arguments": {
    "positions": {"left": null, "right": <normalized opening>},
    "note": "<fresh evidence for grasping or releasing>"
  }
}
\end{promptformat}
\end{minipage}\par
Opening is 0 fully closed and 1 fully open. Results distinguish requested, submitted, and measured values. After detected obstruction, the gripper reference is brought near measured position to avoid accumulating large error. Arms retain their last submitted targets; an unselected side holds.
\end{promptcard}

\begin{promptcard}{F6b\quad Geometric checks and terminal requests}{ALTERNATIVE TOOL SELECTIONS}
\promptpart{Check a path without executing it}
\noindent\begin{minipage}{\linewidth}
\begin{promptformat}
{
  "name": "check_path",
  "arguments": {
    "poses": [
      {"left": <complete pose or null>, "right": <complete pose or null>}
    ],
    "note": "<uncertain approach or permitted orientation to test>"
  }
}
\end{promptformat}
\end{minipage}\par
Check full-path IK, joint limits, and timing without sending arm or gripper commands. Acceptance does not certify collision clearance or physical tracking; execution replans from fresh feedback.

\promptpart{Query image geometry}
\noindent\begin{minipage}{\linewidth}
\begin{promptformat}
{
  "name": "locate_point",
  "arguments": {
    "camera": "<left, right, or top>",
    "pixel_xy": [<image x>, <image y>],
    "reference_step": <earlier observation index>,
    "reference_pixel_xy": [<same feature's earlier x>, <earlier y>],
    "note": "<feature identity and reason for geometric query>"
  }
}
\end{promptformat}
\end{minipage}\par
Pixel origin is the image's top-left corner. The reference fields are optional; a single view supplies a camera ray, not depth. Two wrist views require the same stationary feature and sufficient parallax. Use a metric estimate only when metric\_position\_available=true. A triangulation candidate from degenerate geometry is diagnostic, not a valid motion target.

\promptpart{Finish or report no safe continuation}
\noindent\begin{minipage}{\linewidth}
\begin{promptformat}
{"name": "done",
 "arguments": {
   "summary": "<direct evidence of the physical goal and any uncertainty>",
   "hindsight": "<lessons from the attempt>"
 }}

{"name": "give_up",
 "arguments": {
   "reason": "<attempted strategies and evidence ruling out safe progress>",
   "hindsight": "<lessons from the attempt>"
 }}
\end{promptformat}
\end{minipage}\par
These are alternative responses, not multiple selections in one turn. Completion requires \promptcue{current physical evidence}; the tool itself does not assign the evaluation success label.
\end{promptcard}

\begin{promptcard}[unbreakable]{F7\quad Feedback returned after a request}{NEW INPUT TO THE POLICY}
\promptpart{Selected feedback fields and their roles}
\noindent\begin{minipage}{\linewidth}
\begin{promptformat}
previous_result
  <tool result or concrete rejection reason>
  result / execution_feedback (when returned)
    <requested-versus-measured endpoint errors>
    <per-arm execution and settling diagnostics>

state
  left / right
    joint_pos, joint_vel
    tcp_pose_xyzrpy, tcp_pose_xyzquat
    gripper_normalized, gripper_command_normalized

<fresh wrist and overhead image blocks>
\end{promptformat}
\end{minipage}\par
This is a field-role schematic: result wrapping and diagnostics depend on the tool. An accepted motion plan describes a request; fresh state and execution feedback establish whether it was tracked. Images establish whether the object interaction occurred.

\promptpart{Decision context}
Read a rejection before choosing another target. After execution, compare requested and measured pose, gripper command and measured opening, and visible object motion. Settling diagnostics concern joint stability; they do not by themselves establish accurate TCP arrival, secure grasping, or task completion.
\end{promptcard}

\begin{promptcard}[unbreakable]{P4\quad Recovery and termination instructions}{NEXT-DECISION POLICY}
\promptpart{Recover while preserving the task}
One failed action, tool rejection, missed grasp, occluded target, or uncertain result does not establish impossibility. Diagnose from fresh images, measured state, and previous\_result. Try safe alternatives in viewpoint, approach, grasp, orientation, path, or step size and verify each result.

After IK rejection, preserve required tool-axis directions and compare approach positions, heights, and permitted axial rotations. Use check\_path for uncertain alternatives; do not tilt the gripper merely to make IK pass. The configured workspace is not a measured reachability boundary: an in-range point may still be unreachable at the required orientation. IK acceptance does not certify clearance between camera housings, arms, or objects.

\promptpart{Verify the physical goal}
Call done only when fresh observations establish the requested physical outcome. In the summary, give direct evidence distinguishing completion from an unfinished state, plus any remaining uncertainty. A human assigns the final success/failure label; done itself is not that label.

For insertion, confirm actual mating and seating, not merely resting on the socket. Do not declare completion after withdrawing due to load while the result remains unconfirmed. Continue safe verification or recovery while the outcome is uncertain; do not stop early merely to hand the decision to a human.

\promptpart{When no safe continuation remains}
Do not call give\_up while reasonable safe strategies remain. Consider meaningfully different recoveries. Use give\_up only when evidence shows that the task cannot be completed or further attempts would violate safety constraints. State the attempted strategies and the evidence preventing further progress.
\end{promptcard}

\subsection{Representative Task Prompts}

The examples below retain the task-level wording, with asset references replaced by semantic placeholders. They are supplied together with the applicable shared instructions, context, and tool definitions above, rather than used as standalone one-sentence controller prompts. Prepared task specifications and recorded variants are distinguished where relevant.

\begin{promptcard}[unbreakable]{Cases 1--2\quad Human demonstration tasks}{HUMAN VIDEO / F2}
\promptpart{Pick Red Towel: demonstration-conditioned task}
Watch the historical human demonstration frames and imitate the demonstrated grasping method to pick up the red towel with the robot's right hand in the current scene.

\promptpart{Pick Red Towel: prepared goal-only comparison}
Pick up the red towel with the robot's right hand.

\promptpart{Pick Red Towel: recorded procedural variant without video}
Block the towel with one hand, slide the other hand underneath it, and then grip the towel to lift it.

\promptpart{Pick Up Notebook: recorded task}
Watch the reference video at <notebook demonstration> and pick up the notebook.

\promptpart{Associated context}
The video-conditioned tasks include chronological demonstration image blocks as in F2. The no-video variants omit those blocks. The procedural towel variant provides a grasping method in text; it is not the same instruction as the prepared goal-only comparison.
\end{promptcard}

\begin{promptcard}[unbreakable]{Case 3\quad Unscrew Bottle Cap}{ROBOT VIDEO OR VIDEO + ACTION / F2--F3}
\promptpart{Recorded goal without demonstration}
In the current scene, unscrew and remove the cap from the bottle on the table, and leave the bottle standing securely on the table.

\promptpart{Recorded demonstration-conditioned instruction}
Use the top-view robot demonstration keyframes and corresponding robot states, end-effector poses, and action trajectories in <bottle demonstration> as a reference, adapting the actions to the current observations. In the current scene, unscrew and remove the cap from the bottle on the table, and leave the bottle standing securely on the table.

\promptpart{Mode-specific reference interpretation}
The recorded task wording mentions numerical references in both conditions. In Video, P2 explicitly restricts the supplied reference to images and annotations. In Video + Action, F3 additionally supplies keyframe states and action samples. Numerical references are therefore available only in Video + Action, not inferred from the task wording.

\promptpart{Shared instruction applied to this reference}
Preserve the demonstrated contact side, object-to-gripper orientation, push/pull direction, arm roles, and stage order. Identify the relevant stage in each motion note and explain necessary deviations using live evidence. A demonstration's outcome describes that historical episode only; \promptcue{verify the requested physical result} in the current observations.
\end{promptcard}

\begin{promptcard}[unbreakable]{Case 4\quad Remove and Reinsert Plug}{ROBOT VIDEO OR VIDEO + ACTION / F2--F3}
\promptpart{Recorded Video instruction}
Use the chronological three-view robot demonstration keyframes in <plug video demonstration> as a reference. In the current scene, remove the plug from the power strip on the table, then insert the plug back into the same socket, and leave it fully seated after releasing the gripper.

\promptpart{Recorded Video + Action instruction}
Use the three-view robot demonstration keyframes and corresponding robot states, end-effector poses, and action trajectories in <plug action demonstration> as a reference, adapting the actions to the current observations. In the current scene, remove the plug from the power strip on the table, then insert the plug back into the same socket, and leave it fully seated after releasing the gripper.

\promptpart{Additional control guidance: arm choice and clearance}
Choose the arm or arms using fresh views while preserving the demonstrated roles. Use one arm when sufficient, or both when support is needed and clearance allows it. Explain the arm choice in the first motion note. Keep an unused side null so that it retains its acknowledged target. Check camera, wrist, table, and inter-arm clearance; IK feasibility alone does not establish clearance.

\promptpart{Additional control guidance: orientation and completion}
For a position-only adjustment, hold the acknowledged target quaternion rather than replacing it with measured drift under load. Preserve the required tool direction when comparing approach heights or permitted axial rotations after rejection. When the task requires a table-perpendicular gripper, align tool +z with the table's downward normal; allow axial rotation only when jaw and contact alignment permit it. Without table calibration, base -z is only a proxy. Verify alignment from the measured TCP pose before advancing contact.

Call set\_gripper separately after observing the approach. Before release, verify reliable support; observe finger disengagement before withdrawal. Do not equate a requested pose or gripper state with the physical result. Confirm actual insertion and seating after release, rather than a plug merely resting on the socket.
\end{promptcard}

\begin{promptcard}[unbreakable]{Cases 5--6\quad Goal-image arrangement tasks}{TARGET IMAGE / F4}
\promptpart{Arrange T Shape}
Observe the target image and arrange the blocks on the table into the T shape shown in the image. Match the target image as closely as possible in color, relative position, and spacing.

\promptpart{Arrange Fruit}
Observe the reference image and arrange the four fruits on the table to match the layout shown in the image. Match the target as closely as possible in fruit identity, relative position, and spacing.

\promptpart{Associated input and shared verification}
Each task is paired with its target image and fresh live observations. Shared release instructions require support before opening the gripper and observation before withdrawal. Shared termination instructions require direct evidence of the requested arrangement, rather than merely completing a sequence of placements.
\end{promptcard}

\begin{promptcard}[unbreakable]{Cases 7--8\quad Exploration with self-history}{SELF-HISTORY / F5}
\promptpart{Lemon to Pink Plate: prepared task}
Explore the current scene, locate the pink plate, and place the lemon from the table onto the plate.

\promptpart{Movable Exploration: task specification}
Use the movable exploration history as context. Continue searching for the Sprite bottle on the table by changing the observation position or safely moving obstacles, then grasping it and placing it into the yellow basket containing the strawberry toy once the target is found.

\promptpart{Associated input}
Retained observations, tool requests, and outcomes precede the current observation. This context records what earlier viewpoints revealed and what previous interactions changed. The live state remains the basis for selecting the next tool target.

\promptpart{Shared recovery instruction}
An occluded target or uncertain result does not establish impossibility. Diagnose from fresh images, measured state, and previous tool feedback, then try safe alternatives in viewpoint, approach, grasp, orientation, path, or step size and verify each result.
\end{promptcard}

\begin{promptcard}[unbreakable]{Cases 9--10\quad Online human interaction}{LIVE HUMAN CUES AND INTERACTION HISTORY / F5}
\promptpart{Pointed Fruit Pickup: recorded instruction}
Repeatedly pick up whichever fruit the human points to and place it onto the plate. If no gesture has been made yet, wait for the human's gesture. \promptcue{Stop only when the human makes an OK hand gesture}; otherwise, keep waiting for the human's gesture.

\promptpart{Tic-Tac-Toe: recorded role and turn-taking instruction}
Play a legal game of tic-tac-toe as Green, moving first against an on-site human who plays Blue. Continuously use [Current Board and Human Move History] as context, update the board after every turn, and move only when it is Green's turn and the human's hand has left the board.

\promptpart{Tic-Tac-Toe: move priorities}
Prioritize an immediate win, then block an immediate opponent win, then choose a move that creates a double threat or at least preserves a draw.

\promptpart{Associated input}
These tasks use the live gesture or board, retained interaction history, and fresh robot observations. The fruit instruction defines a continuing interaction with an explicit stop gesture. The game instruction separates legal turn-taking from move selection; a favorable board position does not override the requirement to wait for the human hand to leave.
\end{promptcard}

\par
\endgroup

\end{document}